\pdfoutput=1

\documentclass[11pt]{article}

\usepackage{ACL2023}
\usepackage{graphicx}
\usepackage{booktabs} 
\usepackage{threeparttable} 
\usepackage{multirow}
\usepackage{amssymb} 
\usepackage{bm}
\usepackage{array}
\usepackage{times}
\usepackage{makecell}
\usepackage{latexsym}
\usepackage{graphicx} 
\usepackage{algorithm}
\usepackage{algpseudocode}
\usepackage{amsmath} 
\usepackage[utf8]{inputenc}
\usepackage{graphicx}
\usepackage{booktabs}
\usepackage{tabularx}
\usepackage{amssymb}
\usepackage{xcolor}
\usepackage{xspace}
\usepackage[T1]{fontenc}

\usepackage[utf8]{inputenc}

\usepackage{microtype}

\usepackage{inconsolata}

\newcommand{\eg}{\emph{e.g.},\xspace}
\newcommand{\ie}{\emph{i.e.},\xspace}
\newcommand{\etc}{etc.\xspace}

\newcommand\figref[1]{Fig.~\ref{#1}}

\newcommand\tabref[1]{Table~\ref{#1}}

\newcommand\secref[1]{Sec.~\ref{#1}}

\newcommand{\fakeparagraph}[1]{\vspace{1mm}\noindent\textbf{#1.}}

\newcommand{\modelname}{{MGT-Prism}\xspace}
\title{\textit{MGT-Prism}: Enhancing Domain Generalization for Machine-Generated Text Detection via Spectral Alignment}

\author{Shengchao Liu\textsuperscript{$1$}, Xiaoming Liu\textsuperscript{$1,\ast$}, 
Chengzhengxu Li\textsuperscript{$1$},  
\\\textbf{Zhaohan Zhang\textsuperscript{$2$},
Guoxin Ma\textsuperscript{$1$},  
Yu Lan\textsuperscript{$1$},  
Shuai Xiao\textsuperscript{$3$}} \\
        \textsuperscript{1}Faculty of Electronic and Information Engineering, Xi'an Jiaotong University\\ 
        \textsuperscript{2}Queen Mary University of London,
        \textsuperscript{3}Alibaba  \\
        \texttt{
        \{liusc, czx.li, guoxin.ma\}@stu.xjtu.edu.cn, zhaohan.zhang@qmul.ac.uk
        }\\
        \texttt{
        \{xm.liu, ylan2020\}@xjtu.edu.cn, shuai.xsh@alibaba-inc.com
        }\\
        }

\begin{document}
\maketitle
\renewcommand{\thefootnote}{\fnsymbol{footnote}}
\footnotetext[1]{Corresponding author}
\renewcommand{\thefootnote}{\arabic{footnote}}
\begin{abstract}
Large Language Models have shown growing ability to generate fluent and coherent texts that are highly similar to the writing style of humans.
Current detectors for Machine-Generated Text (MGT) perform well when they are trained and tested in the same domain but generalize poorly to unseen domains, due to domain shift between data from different sources.
In this work, we propose \modelname, an MGT detection method from the perspective of the frequency domain for better domain generalization.
Our key insight stems from analyzing text representations in the frequency domain, where we observe consistent spectral patterns across diverse domains, while significant discrepancies in magnitude emerge between MGT and human-written texts (HWTs).
The observation initiates the design of a low frequency domain filtering module for filtering out the document-level features that are sensitive to domain shift, and a dynamic spectrum alignment strategy to extract the task-specific and domain-invariant features for improving the detector's performance in domain generalization.
Extensive experiments demonstrate that \modelname\ outperforms state‑of‑the‑art baselines by an average of 0.90\% in accuracy and 0.92\% in F1 score on 11 test datasets across three domain‑generalization scenarios.
\end{abstract}

\section{Introduction}

Large Language Models (LLMs) are becoming popular as writing assistants in daily work for their incredible ability to generate fluent and coherent texts following users' instructions.
However, the widespread applications of LLMs have raised substantial concerns regarding their misuse in fake news generation \cite{liu2024does}, ghostwriting \cite{kumar-etal-2025-mixrevdetect}, spamming, \etc \cite{wang-etal-2024-stumbling,li2025ironsharpensirondefending,zellers2019defending,pudasaini-etal-2025-benchmarking}, which calls for an urgent need to detect MGTs precisely and reliably.

\begin{figure}[t]
  \centering
  \includegraphics[width=\columnwidth]{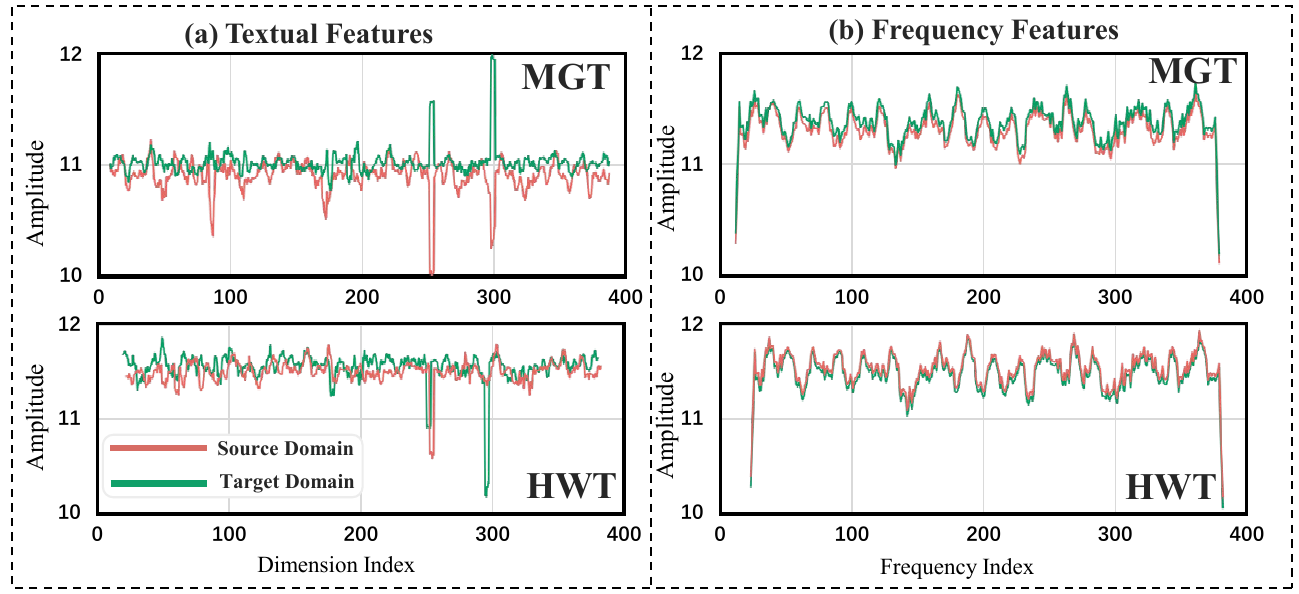}
  \caption{ 
    \textbf{Comparison of the textual features and their corresponding frequency features.} 
The horizontal axis represents the feature dimension index for both textual and frequency-domain features, while the vertical axis shows the magnitude of raw feature values in the text domain and transformed frequency components, respectively.
    Frequency features exhibit more consistent patterns between intra-class samples from different domains, providing the foundation for frequency-based detection under domain shift\protect\footnotemark.
    }
  \label{fig:mae} 
\vspace{-0.2cm}
\end{figure}

\footnotetext{The data used in our experiments consist of 3,000 randomly sampled HWTs and 3,000 MGTs from both the source and target domains, covering all three datasets used in domain generalization.}

Due to the diversity in application scenarios, model architectures, and scales of LLMs, an ideal MGT detector should perform consistently well in \emph{domain generalization} \footnote{In this work, we use \emph{domain generalization} to refer to generalization across MGT domains, MGT generators, and generator scales in MGT detection.}(DG)  settings where the detector is trained on datasets from different source domains and tested on unseen target domains.  
Two mainstream approaches dominate current MGT detection methods, \ie fine-tuning and metrics-based methods, yet both largely neglect the detector's ability to generalize.
Existing fine-tuning methods either focus on capturing distinguishing features between MGT and HWT, such as text coherence~\cite{liu-etal-2023-coco}, token probabilities~\cite{chen2025imitate,bao2023fast,wang-etal-2023-seqxgpt}, and attention patterns~\cite{kushnareva2021artificial}, or explore novel training strategies, including contrastive learning~\cite{liu-etal-2024-detectgpt}, and adversarial training~\cite{hu2023radar, li2025ironsharpensirondefending}.  
However, these methods do not explicitly disentangle task-specific and domain-specific features during training, resulting in limited domain generalization.
Metric-based methods work in an unsupervised manner.
They calculate a single score for distinguishing MGTs from HWTs using perplexity \cite{bao2024fastdetectgpt}, entropy \cite{shum2023automatic}, or re-sampling \cite{Shi_2024}.
Metric-based methods also suffer from performance degradation in DG settings, as they rely on a specific training set for deciding the optimized classification threshold.

Inspired by the potential of the frequency domain to disentangle the feature representations into orthogonal components \cite{tamkin2020language,sun-etal-2024-unleashing,guo2023aloft}, we examine the inter- and intra-class features from the perspective of spectra using Discrete Fourier Transform (DFT) \cite{bracewell1986}.
The preliminary analysis in \secref{sec:Preliminary} reveals a critical property of text representations in the frequency domain. As shown in Figure \ref{fig:mae}, the spectrum of intra-class text representation from different domains exhibits differences in the magnitude but is more consistent in the distribution pattern, indicating that \textit{i)} distinguishable features between MGT and HWT exist in the frequency domain; \textit{ii)} transfer from features space to frequency domain mitigates the domain shift.

Based on the natures of spectra, which are decomposability and insensitivity to domain shift, we propose \modelname,  an MGT detection framework from the perspective of the frequency domain, to align intra-class features from different domains for emphasizing the task-specific and domain-invariant inter-class features, therefore enabling strong DG ability of MGT detectors. 
Specifically, we design a frequency spectrum filter block to suppress domain-related features \cite{tamkin2020language}.
To mitigate the domain shift in finer-grain, we further introduce a frequency spectrum alignment strategy to reduce the distribution discrepancy among intra-class instances in the frequency domain.
By aligning intra-class features in the frequency domain, \modelname successfully extracts and utilizes the task-specific and domain-invariant features from the training set and generalizes well to unseen domains.
Extensive experiments demonstrate the strong generalization ability of \modelname among multiple settings.
Compared to state-of-the-art approaches in accuracy and F1 score, \modelname{} achieves average improvements of 0.92\% and 1.56\%, 0.90\% and 0.69\%, and 0.88\% and 1.24\% in cross-generator, cross-domain, and cross-scale, respectively, highlighting its efficacy.
Our contributions are summarized as follows:

\begin{itemize}
    \item We propose to analyze MGT detection in the frequency domain and observe a similar distribution pattern with varied magnitudes between MGT and HWT spectra, offering a new insight and broad applicability in MGT detection.
    
    \item We design a novel model, \modelname, which aligns the intra-class features in text spectra from different domains to extract task-specific and domain-invariant features for enhancing domain generalizability in MGT detection.

    \item We conduct extensive experiments on 11 test sets across three generalization scenarios. The results demonstrate that our model consistently outperforms state-of-the-art methods in both effectiveness and generalization capability for MGT detection. 
    
\end{itemize}

\section{Related Work}

\fakeparagraph{Domain Generalization} 
In the context of MGT detection, DG means a detector should remain reliable when faced with text from new generators or domains not encountered during training.
Many metric-based detectors rely on head-token analysis \cite{gehrmann2019gltr}, logistic regression on perplexity features \cite{bao2024fastdetectgpt,bao2023fast}, or token-wise log probabilities \cite{wang2023seqxgpt,hans2024spotting} from white-box LLMs to address this challenge.
While metric-based methods are training-free, their prediction accuracy is often lower than supervised approaches. In contrast, fine-tuned detectors benefit from perturbation-based methods \cite{li2025learningrewritegeneralizedllmgenerated,gao2023mask,shum2023automatic}, contrastive learning \cite{tack2020csi,gunel2020supervised}, or a combination of both \cite{liu2024does}, which have proven effective in improving model generalization. In addition, adversarial training frameworks \cite{li2025ironsharpensirondefending,hu2023radar} have been employed to address robustness issues. 
Unlike prior methods based on overall text features, our approach introduces the Fourier Transform to extract multi-scale frequency information, emphasizing both global and local patterns, and mitigates distributional bias to enable effective detection in DG.

\fakeparagraph{Fourier Transforms}
Recently, frequency-based techniques have been increasingly integrated into the computer vision field \cite{yue2025freeformerfrequencyenhancedtransformer,yi2023frequency,guo2023aloft}. By analyzing how low frequencies in an image typically capture global structures \cite{fan2022depts} and color information \cite{cao2020spectral}, while high frequencies contain fine details of objects, a series of methods to enhance predictive capabilities have emerged \cite{woo2022cost,fan2022depts}. Some works have applied these techniques to the natural language processing (NLP) field \cite{wu2021autoformer,lee2021fnet}, combining them with attention mechanisms to accelerate computations \cite{tamkin2020language,lee2021fnet,choromanski2020masked}, 
and embedding spectral filters in different tasks \cite{NEURIPS2024_cf112959,tamkin2020language,khan2019regularization,rippel2015spectral}, where the low-frequency components are considered to represent slower-changing features.
Inspired by this, we observe that transforming representations from the feature space to the frequency domain reduces dimensional complexity. It also enhances the structural regularity of feature distributions in frequency domain.
This transformation amplifies distributional irregularities (\eg lexical repetition, unnatural conjunctions, and templated phrasing), making MGT and HWT more distinguishable for downstream detection.

\section{Preliminary}\label{sec:Preliminary}

This section briefly introduces the problem formulation and the Discrete Fourier Transform (DFT) in traditional signal processing and its application in our work.
\subsection{MGT Detection under DG}
The DG task is defined as follows: 
given a training set consisting of multiple observed source domains  
\(\mathcal{D}_S=\{\mathcal{D}_n\}_{n=1}^{N}\), 
where each domain 
\(\mathcal{D}_n=\{(x_i^{(n)}, y_i^{(n)})\}_{i=1}^{T_n}\) 
contains \(T_n\) labeled samples, 
the goal is to learn a model from \(\mathcal{D}_S\) that generalizes to arbitrary unseen target domains \(\mathcal{D}_T\) whose distributions differ from those of the sources.

\subsection{Discrete Fourier Transform}
The Continuous Fourier Transform (CFT) is one of the core mathematical tools in the field of signal processing \cite{bracewell1986}. 
The idea is that any complex signal can be decomposed into a superposition of sine waves (or complex exponentials) of different frequencies. 
Specifically, for any time-domain signal $x(t)$, its frequency component $X(f)$ at frequency $f$ in the frequency domain is expressed as:
\begin{align}
    X(f)=\int_{-\infty}^{\infty} x(t)e^{-2\pi jft}dt,
\end{align}
where \(e^{-2\pi jft}\) is the complex exponential basis and \(j\) denotes the imaginary unit. The DFT extends the CFT to discrete signals, where \(x[n]\) denotes uniformly sampled values of \(x(t)\). The \(k\)-th frequency component in the frequency domain is:
\begin{equation}\label{2}
    X[k]=\sum_{n=0}^{N-1} x[n]e^{-2\pi j\frac{kn}{N} }, \quad 0 \leq k \leq (N-1),
\end{equation}
where $N$ represents the length of the signal, which is the number of sampling points of the input sequence $x[n]$.


\subsection{Applying DFT to Contextual Word Representations}\label{ADFT}
 
Given an input sequence $x[n]$, an encoder-only Transformer (\eg RoBERTa) produces final-layer token vectors
\(V=(v_0,\ldots,v_{N-1})\), where \(v_t\in\mathbb{R}^d\).
Fix a neuron \(i\in\{0,\ldots,d-1\}\) and consider its trajectory across tokens
\(s^{(i)}=(v_0[i],\ldots,v_{N-1}[i])\in\mathbb{R}^N\).
Building on Eq.~\eqref{2}, which treats the final-layer \texttt{[CLS]} vector as a length-\(d\) signal along the hidden dimension, we generalize to the token axis: fixing neuron \(i\), we apply the same DFT to its activations across tokens \(s^{(i)}\). The spectral coefficients \(h^{(i)}_k\) are: 
\begin{align}\label{3}
h^{(i)}_k=\sum_{t=0}^{N-1} s^{(i)}_t\,e^{-2\pi \mathrm{i} k t/N},\qquad k=0,\ldots,N-1 .
\end{align}
Here \(k=0\) is the lowest component (Direct Current component).
Low-, high-, or band-pass effects are obtained by applying a spectral mask \(m_k\) (\eg zeroing selected \(h^{(i)}_k\)).

\subsection{Why Frequency Helps MGT Detection Generalize Across Domains}
A common property of modern NLP models is their ability to produce \emph{contextualized} token representations by modeling a sequence of tokens (e.g., characters or subword units).
We collect the frequency-domain features as 
\(\mathbf{H}=\{\mathbf{h}_k\}_{k=0}^{N-1}\), where 
\(\mathbf{h}_k=[h^{(0)}_k,\ldots,h^{(d-1)}_k]^\top\).
To capture global structural patterns, we keep components that complete at most one full oscillation over the input. 
Consequently, we define the low-frequency band as:
\begin{equation}
\mathbf{H}_{\text{low}}=\{\mathbf{h}_{k}\}_{k=0}^{d_{\text{low}}},\quad
d_{\text{low}} = \left\lceil \frac{N}{t_{\text{num}}} \right\rceil.
\label{eq:low}
\end{equation}
where $t_{\text{num}}$ is the number of tokens in the input text.
Similarly, to capture sentence-level features, we retain components with at most one oscillation per sentence, based on the sentence count \(s_{\text{num}}\):
{\small
\begin{equation}
\mathbf{H}_{\text{mid}}=\{\mathbf{h}_{k}\}_{k=d_{\text{low}}+1}^{d_{\text{mid}}},\quad
d_{\text{mid}}=\left\lceil \frac{N}{s_{\text{num}}} +(N-d_{\text{low}})\right\rceil.
\label{eq:medium_none}
\end{equation}
}
After obtaining the mid frequency band $\mathbf{H}_{\text{mid}}$, we divide the remaining frequency features into high frequency band:
\begin{align}
\mathbf{H}_{\text{high}}=\{\mathbf{h}_{k}\}_{k=d_{\text{mid}}+1}^{N-1}.
\label{eq:high}
\end{align}

The inverse DFT (IDFT; \citet{lee2021fnet}), which transforms the input from the frequency domain $\mathbf{H}_{\text{f}}$ back to the text feature space $H_f$, can be expressed as:
\begin{equation}\label{eq:idft0}
    H_{f} = \text{IDFT}(\mathbf{H}_{\text{f}})
\end{equation}



Previous studies \cite{tamkin2020languageprismspectralapproach} have shown that the high-frequency components reflect word-level features (\eg perplexity, log-probability ), while low-frequency components are associated with document-level characteristics (\eg topic, style).
Correspondingly, recent studies DetectGPT\cite{mitchell2023detectgpt} and Binoculars \cite{hans2024spotting} demonstrate that the distinct writing preference at the word-level plays a key role in MGT detection. 
Therefore, we conduct two complementary validations \footnote{Our experimental data are identical to those used in Figure~1.}: \textbf{\textit{i)} in the frequency domain}, we examine whether high-frequency components are particularly sensitive to token-level changes; and \textbf{\textit{ii)} in the feature space} , we assess whether low-frequency components effectively capture global structural features. These findings support the subsequent use of low-frequency filtering and frequency-domain alignment to improve DG in MGT detection.

\fakeparagraph{Analyzing the Correlation Between Frequency Components and Feature Space}
To validate whether different frequency components effectively capture distinct linguistic features, we apply three types of content-preserving perturbations in the feature space, namely token perturbations (word-level), sentence reordering (sentence-level), and theme transformations (document-level).
As shown in Table~\ref{tab:maeshift}, in the feature space, token-level perturbations tend to cause larger MAE\footnote{Mean Absolute Error is a standard metric for measuring signal-level differences, and a higher MAE reflects greater spectral deviation between the perturbed and original inputs in the frequency domain \cite{yi2023frequency}. The perturbation rate is 15\%.
}  values in the high-frequency components, while theme transformations primarily increase the MAE in the low-frequency components compared to other perturbations. This indicates that high-frequency components are more sensitive to token-level changes and predominantly encode token-level features.
Conversely, low-frequency components are more sensitive to document-level perturbations, capturing domain-specific information (\eg topic, style).
Further analysis of domain shift is provided in Appendix~B.3.

\begin{table}[H]
\centering
\resizebox{\columnwidth}{!}{
\begin{tabular}{l|ccc}
\toprule
\textbf{Perturbation} & \textbf{low-frequency} & \textbf{mid-} & \textbf{high-} \\
\midrule

Theme-Transformation & \textbf{0.0072} & 0.0166 & 0.0349   \\
Sentence-Reordering    & 0.0002 & \textbf{0.0744} & 0.0250  \\
\midrule
Token-Replacement    & 0.0041 & 0.0867 & \textbf{0.3225}   \\
Token-Delete    & 0.0034 & 0.2453 & \textbf{0.4122}   \\
Token-Repetition    & 0.0048 & 0.2703 & \textbf{0.5592}  \\

\bottomrule
\end{tabular}
}
\caption{MAE Shift in Frequency Bands.}  
\label{tab:maeshift}
\end{table}

\fakeparagraph{Topic Coherence in Feature Space}  
To evaluate the coherence between frequency components and global semantic features, we project each frequency band back into the feature space via IDFT  and compute its similarity to the original document-level embedding using BERTScore\footnote{BERTScore \cite{angelov2024topic} is a widely used metric for assessing the topic coherence of learned representations. Higher scores indicate stronger global semantic coherence between the reconstructed embeddings and the original embeddings.}
. As shown in Table~\ref{tab:bertscore}, low-frequency components exhibit the highest similarity, whereas high-frequency components score the lowest, indicating that low-frequency features better preserve global topical information.

\begin{table}[H]
\centering
\resizebox{\columnwidth}{!}{
\begin{tabular}{l|ccc}
\toprule
\textbf{Perturbation} & \textbf{low-frequency} & \textbf{mid-} & \textbf{high-} \\
\midrule

BERTSore & \textbf{0.8671} & 0.4062 & 0.1945   \\

\bottomrule
\end{tabular}
}
\caption{Topic Coherence Compared to the Original Text.}
\label{tab:bertscore}
\end{table}

\begin{figure*}
    \centering
    \includegraphics[width=\textwidth]{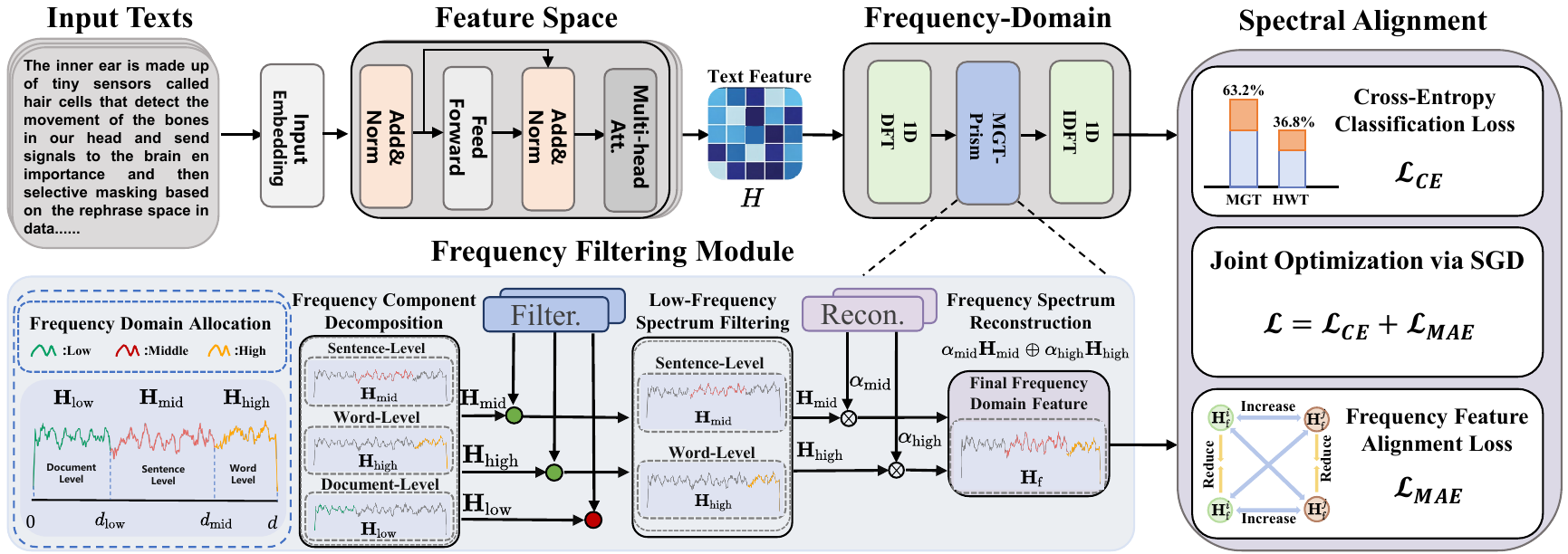}
    \caption{{Overview of \modelname{}.} In the Low Frequency Filtering Module  (\secref{sec:Per}), we transform features from feature space into the frequency domain, where we analyze information distribution across multiple frequency bands (\ie low-, mid-, and high-frequency). Then, we propose a low-frequency filtering module to suppress redundant features shared between MGT and HWT.
    In the Spectral Alignment stage (\secref{sec:mcl}), we compute the global frequency spectrum distribution and reconstruct the frequency components accordingly. Furthermore, we design a frequency alignment loss to enhance DG.}
    \label{fig:duco}
\end{figure*}
\section{Methodology}

The workflow of the proposed \modelname{} is shown in \figref{fig:duco}, comprising two main components: a Low Frequency Filtering Module (LFF), and a Spectrum Alignment Module (SAM), to reduce the domain gap for intra-class samples.
LLF cuts out the low-frequency band to reduce domain-sensitive but task-irrelevant features.
SAM sets an optimization objective to mitigate domain gap for data within the same class.
The learning process of our algorithm is described in Appendix A.3.

\subsection{Low Frequency Filtering Module \label{sec:Per}}

Previous studies \cite{tamkin2020languageprismspectralapproach,sun-etal-2024-unleashing} on frequency domain analysis in the field of natural language processing and analysis about the discrepancy between MGT and HWT \cite{chen2025imitate, liu-etal-2024-detectgpt} observe that the low-frequency band corresponds with document-level features. 
 We transform the text input into a frequency domain representation \cite{lee2021fnet, yi2023frequency} and decompose it into three frequency bands (\ie low, mid, and high), and propose a low-frequency filtering module to suppress the common features between MGT and HWT.

With the improvement of text generation capabilities of LLMs, MGT hardly differs from HWT at the discourse level \cite{chen2025imitate,liu2024does}. 
In frequency domain analysis, this phenomenon is reflected in the fact that low frequency components are often smoother \cite{tamkin2020language}. 
Furthermore, in the DG settings, changes in style, generator, or theme are more likely to cause the migration of text  document-level features. 
Therefore, we filter out low-frequency components $\mathbf{H}_{\text{low}}$ while retaining mid and high frequency components to enhance the classifier's ability to distinguish between MGT and HWT under DG.
The final frequency domain features after low-frequency filtering can be expressed as:
\begin{align}
d_{\text{mid}}=\left\lceil \frac{N}{s_{\text{num}}} \cdot \tau +(N-d_{\text{low}})\cdot(1-\tau)\right\rceil,
\label{eq:medium2}
\end{align}
\begin{equation}
\mathbf{H}_{\text{mid}}=\{\mathbf{h}_{k}\}_{k=d_{\text{low}}+1}^{d_{\text{mid}}},\quad
    \mathbf{H}_{\text{f}} = \mathbf{H}_{\text{mid}}  \oplus \mathbf{H}_{\text{high}},
\label{eq:high10}
\end{equation}
where $\tau \in \left[0,1\right]$ is the scaling factor.
By adjusting $\tau$, the mid-frequency band will be affected by the number of sentences and the remaining frequency bands, avoiding the overflow problem caused by too few or too many sentences.
$\oplus$ denote the concatenation operation. 
From the perspective of natural text, $\mathbf{H}_{\text{f}}$ retains more fine-grained information of words and sentences in the text input $x$.

\subsection{Spectral Alignment Module} \label{sec:mcl}
\fakeparagraph{Frequency Spectrum Reconstruction} After removing the low-frequency features that are easily affected by domain-shift, we restore the remaining frequency-domain features to the original features to reduce the detector's excessive attention to specific domain information during the training process.
Specifically, given a batch of training data $D=\{x_{b},y_{b}\}_{b=0}^{B-1}$ with $B$ text-label pairs.
Through the previous section, we can obtain the frequency domain features representation $\{\mathbf{H}_{\text{f}}^{b}\}_{b=0}^{B-1} = \{\mathbf{H}_{\text{mid}}^{b} \oplus \mathbf{H}_{\text{high}}^{b}\}_{b=0}^{B-1}$ of each input text $T_{b}$.
Subsequently, we compute the average modulus values of the frequency-domain features within the mid- and high-frequency intervals for the current batch $D$.
These values are then compared with the corresponding average modulus values of the entire training dataset to determine the weighting factors $\alpha _{\text{mid}}$ and $\alpha _{\text{high}}$ for reconstructing the features in the mid and high frequency bands.
\begin{equation}
\mu_{\text{mid}} = \frac{1}{B}\sum_{b=0}^{B-1}  \left | \mathbf{H}_{\text{mid}}^{b} \right |, \enspace \mu_{\text{high}} = \frac{1}{B}\sum_{n=0}^{B-1}  \left | \mathbf{H}_{\text{high}}^{b} \right |,
\end{equation}
\begin{equation}
\alpha _{\text{mid}}=\frac{\bar{\mu}_{\text{mid}} }{\mu_{\text{mid}}}, \enspace \alpha _{\text{high}}=\frac{\bar{\mu}_{\text{high}} }{\mu_{\text{high}}} ,
\end{equation}
where $\left |.\right |$ represents the modulus operation, $\bar{\mu}_{\text{mid}}$ and $\bar{\mu}_{\text{high}}$ represent the average modulus values of the mid and high frequency of the entire training set, respectively.
Therefore, the final frequency domain features can be expressed as:
\begin{equation}
\mathbf{H}_{\text{f}}^{b} = \alpha _{\text{mid}}\mathbf{H}_{\text{mid}}^{b} \oplus \alpha _{\text{high}}\mathbf{H}_{\text{high}}^{b},\enspace 0 \leq b \leq (B-1).
\label{eq:13}
\end{equation}
The natural text features after filtering and reconstruction can be expressed as:
\begin{equation}\label{eq:14}
    H_{f}^{b} = \text{IDFT}(\mathbf{H}_{\text{f}}^{b}),\enspace 0 \leq b \leq (B-1),
\end{equation}
By low frequency filtering and reconstructing frequency domain features, we strip away the global information that is less discriminative between MGT and HWT, thereby promoting more effective learning of local features.

\fakeparagraph{Frequency Spectrum Alignment} 
To further promote feature alignment among samples from the same class and enhance generalization to unseen domains, we introduce a frequency domain feature alignment loss \(\mathcal{L}_{\text{MAE}}\).
Building on the standard signal level dissimilarity metric, mean absolute error (MAE) \cite{yi2023frequency}, \(\mathcal{L}_{\text{MAE}}\) reduces distributional differences by minimizing the average \(L_{1}\) distance~\cite{zhou2021contrastive} between samples from the same class in the frequency domain.

Specifically, given a batch of training data $D$
and corresponding frequency features $\{\mathbf{H}_{\text{f}}^{b}\}_{b=0}^{B-1}$, for any frequency features $\mathbf{H}_{\text{f}}^{b}$, let the set of remaining frequency features with the same label as $D_{\text{pos}}$, and the set of remaining frequency features with different labels be denoted as $D_{\text{neg}}$.
Then the loss function $\mathcal{L}_{\text{MAE}}$ can be defined as:
\begin{equation}
\mathcal{L}_{\text{pos}} = \mathbb{E}_{\mathbf{H}_{\text{f}}^{b}\in D} \mathbb{E}_{\mathbf{H}_{\text{f}}^{i}\in D_{\text{pos}}}(\| (|\mathbf{H}_{\text{f}}^{b}| - |\mathbf{H}_{\text{f}}^{i}|) \|_1),
\end{equation}
{\small
\begin{equation}
\mathcal{L}_{\text{neg}} = \mathbb{E}_{\mathbf{H}_{\text{f}}^{b}\in D} \mathbb{E}_{\mathbf{H}_{\text{f}}^{i}\in D_{\text{neg}}}(\max (0, ( \xi - \| ( |\mathbf{H}_{\text{f}}^{b}| - |\mathbf{H}_{\text{f}}^{i}|) \|_1 ))),
\end{equation}
}

\begin{equation}\label{eq:18}
\mathcal{L}_\text{MAE} = \mathcal{L}_{\text{pos}} + \mathcal{L}_{\text{neg}},
\end{equation}
where $\|\cdot\|_1$ denotes the MAE computed over the
modulus values, $\xi$ represents the maximum same class distance, which is used to control the distance between samples with different labels. Then, the total loss is computed by:
\begin{equation}
    \mathcal{L} = \mathcal{L}_{\text{CE}} + \mathcal{L}_{\mathrm{MAE}},
\end{equation}
where the cross-entropy classification loss $\mathcal{L}_{\text{CE}}$ is computed on natural text features $H_{f}^{b}$, describing in Eq. \eqref{eq:14}, and corresponding labels $Y_{b}$.

\begin{table*}[ht]
\centering
\resizebox{\textwidth}{!}{
\begin{tabular}{ccc|cccc|ccccc|c}
\toprule
\multicolumn{3}{c|}{\textbf{Method}} 
& \multicolumn{4}{c|}{\textbf{Metric-based}} 
& \multicolumn{5}{c}{\textbf{Model-based}} \\
\midrule
\textbf{Dataset} 
& \textbf{Test data}
&Metric
& \textbf{\textit{GLTR}} 
& \textbf{\textit{DetectGPT}} 
& \textbf{\textit{Fast-Dete.}}
& \textbf{\textit{Binoculars}}
& {\textbf{\textit{RoBERTa}$_{\textbf{}}^\dag$}} 
& \textbf{\textit{RADAR}} 
& \textbf{\textit{Ghostbuster}} 
& {\textbf{\textit{PECOLA}$_{\textbf{}}^\dag$}} 
& \textbf{\textit{ImBD}} 
& {\textbf{\textit{\modelname{}}$_{\textbf{}}^\dag$}} \\
\toprule

\multirow{8}{*}{\textbf{\rotatebox{90}{\shortstack{Cross--Generator}}}}
& \multirow{2}{*}{FLAN-T5}   & Acc & 67.10 & 60.30 & 74.70&63.46 &\underline{86.80$_{\text{1.90}}$} & {65.82$_{\text{4.06}}$} & 79.55$_{\text{4.16}}$ & 86.30$_{\text{1.80}}$&   71.55$_{\text{2.06}}$ & \textbf{89.62$_{\textbf{1.67}}$} \\

&   & F1 & 52.11 & 49.33 & 60.70& 52.70&{85.95$_{\text{1.76}}$} & {60.75$_{\text{5.06}}$} & 75.45$_{\text{4.26}}$ & \underline{86.20$_{\text{1.62}}$} & 65.54$_{\text{3.46}}$ & \textbf{89.02$_{\textbf{1.97}}$} \\

& \multirow{2}{*}{ChatGPT}&Acc   & 78.12 & 60.71 & 76.60&\underline{87.60} & 85.01$_{\text{3.74}}$ & 61.50$_{\text{3.01}}$ & 77.06$_{\text{2.04}}$ & 84.21$_{\text{3.50}}$&  {85.68$_{\text{3.76}}$} & \textbf{88.96$_{\textbf{2.03}}$} \\
&   & F1 & 62.01 & 68.70 & 69.70& \underline{87.63} &{84.65$_{\text{1.76}}$} & {69.95$_{\text{5.15}}$} & 72.65$_{\text{5.07}}$ & 84.25$_{\text{2.01}}$ &79.55$_{\text{3.26}}$ & \textbf{89.22$_{\textbf{1.67}}$} \\

& \multirow{2}{*}{GLM}
& Acc      & 75.50 & 73.20 & 76.02& \textbf{95.28} & 89.94$_{\text{1.68}}$ & 70.04$_{\text{4.09}}$ & 80.94$_{\text{2.58}}$ & {89.60$_{\text{2.70}}$} & 87.48$_{\text{2.06}}$ & \underline{92.42$_{\textbf{1.56}}$} \\
&   & F1 & 60.10 & 62.43 & 69.62&\textbf{95.36} &{86.95$_{\text{2.06}}$} & {62.90$_{\text{3.09}}$} & 74.65$_{\text{4.13}}$ & 84.22$_{\text{2.01}}$&  89.04$_{\text{3.76}}$ & \underline{92.82$_{\textbf{1.67}}$} \\

& \multirow{2}{*}{LLaMA}&Acc     & 72.40 & 70.80 & 79.65& 80.95 & \underline{85.31$_{\text{2.67}}$} & 70.12$_{\text{4.67}}$ & 80.45$_{\text{2.76}}$ & 82.99$_{\text{2.61}}$ & 84.22$_{\text{3.07}}$ & \textbf{87.66$_{\textbf{1.74}}$} \\
&   & F1 & 60.17 & 62.33 & 61.60& 83.15 &{80.85$_{\text{3.66}}$} & {67.95$_{\text{3.05}}$} & 79.65$_{\text{4.01}}$ & 84.20$_{\text{1.92}}$&  \underline{85.34$_{\text{2.06}}$} & \textbf{87.20$_{\textbf{1.77}}$} \\

\midrule

\multirow{8}{*}{\textbf{\rotatebox{90}{Cross--Domain}}}
& \multirow{2}{*}{Opinion.S}&Acc    & 71.15 & 70.40 & 76.30& \underline{95.04} & 93.61$_{\text{2.65}}$ & 74.39$_{\text{3.05}}$ & 90.97$_{\text{2.91}}$ & 93.49$_{\text{3.06}}$ & {94.85$_{\text{0.33}}$} & \textbf{96.06$_{\textbf{1.08}}$ }\\
& &F1 & 65.60 & 60.70 & 72.70&\underline{95.20} & {92.24$_{\text{1.93}}$} & {60.04$_{{2.56}}$} & 91.27$_{\text{1.83}}$ & 91.14$_{\text{2.20}}$ & 92.58$_{\text{2.06}}$ & \textbf{96.90$_{\textbf{1.53}}$} \\

& \multirow{2}{*}{Question.A}&Acc       & 72.90 & 72.20 & 76.90 & 93.60 & \underline{93.92$_{\text{1.15}}$} & 72.52$_{\text{1.95}}$ & 92.82$_{\text{4.06}}$ & 94.12$_{\text{2.95}}$ & 93.23$_{\text{0.92}}$ & \textbf{96.53$_{\textbf{1.45}}$} \\
& &F1 & 70.60 & 76.10 & 70.80&\underline{93.62} & {93.04$_{\text{1.93}}$} & {78.64$_{{3.64}}$} & 91.45$_{\text{2.74}}$ & 92.64$_{\text{2.10}}$&  92.28$_{\text{1.06}}$ & \textbf{96.20$_{\textbf{1.13}}$} \\

& \multirow{2}{*}{Scientific.W}&Acc  & 41.24 & 50.90 & 66.90& \underline{79.92} & 70.43$_{\text{8.29}}$ & 70.18$_{\text{10.79}}$ & 62.73$_{\text{9.19}}$ & 68.90$_{\text{9.17}}$ & \textbf{80.26$_{\textbf{5.10}}$} & {79.88$_{\text{7.22}}$} \\
& &F1 & 32.60 & 35.22 & 48.70& \textbf{79.95} & {61.94$_{\text{8.23}}$} & {50.64$_{{8.94}}$} & 55.95$_{\text{6.93}}$ & 60.25$_{\text{7.80}}$&  74.28$_{\text{4.56}}$ & \underline{77.80$_{\textbf{7.43}}$} \\

& \multirow{2}{*}{Story.G}&Acc     & 70.60 & 60.22 & 76.70&90.62 & \underline{98.62$_{\text{0.42}}$} & 75.76$_{\text{3.23}}$ & 95.60$_{\text{1.32}}$ & 98.20$_{\text{0.40}}$&  92.75$_{\text{0.54}}$ & \textbf{98.96$_{\textbf{0.25}}$} \\
& &F1 & 65.25 & 67.06 & 70.90 &90.90 & \underline{98.04$_{\text{1.63}}$} & {70.84$_{{3.06}}$} & 94.65$_{\text{2.63}}$ & 97.09$_{\text{2.00}}$&  93.08$_{\text{2.76}}$ & \textbf{98.90$_{\textbf{1.03}}$} \\

\midrule

\multirow{6}{*}{\textbf{\rotatebox{90}{\shortstack{Cross--Scale}}}}
& \multirow{2}{*}{LLaMa-13b}&Acc & 73.92 & 71.20 & 82.90&94.65 & 93.85$_{\text{1.72}}$ & 76.28$_{\text{2.60}}$ & 90.95$_{\text{1.92}}$ & 93.48$_{\text{2.12}}$&  \underline{94.91$_{\text{0.95}}$} & \textbf{95.94$_{\textbf{1.62}}$} \\
& &F1 & 70.60 & 75.02 & 78.98& \underline{93.70} & {92.74$_{\text{1.50}}$} & {70.60$_{{3.14}}$} & 92.50$_{\text{1.93}}$ & 92.14$_{\text{2.01}}$&  92.78$_{\text{2.06}}$ & \textbf{95.91$_{\textbf{1.70}}$} \\

& \multirow{2}{*}{LLaMa-30b}&Acc & 77.90 & 73.20 & 80.62& 93.65 & 93.28$_{\text{1.20}}$ & {77.49$_{\text{1.25}}$} & 94.44$_{\text{2.10}}$ & \underline{94.44$_{\text{1.19}}$}  & 92.26$_{\text{1.71}}$ & \textbf{94.67$_{\textbf{1.20}}$} \\
& &F1 & 73.90 & 77.25 & 84.21 & \underline{94.02} & {92.60$_{\text{1.93}}$} & {71.64$_{{3.14}}$} & 92.15$_{\text{2.76}}$ & 93.04$_{\text{1.40}}$& 90.28$_{\text{3.66}}$ & \textbf{93.70$_{\textbf{1.73}}$} \\

& \multirow{2}{*}{LLaMa-65b}&Acc & 75.60 & 74.90 & 77.92&{92.05} &  \underline{92.24$_{\text{1.63}}$} & {74.70$_{{2.60}}$} & 91.95$_{\text{2.03}}$ & 92.06$_{\text{1.10}}$&  86.38$_{\text{2.90}}$ & \textbf{93.60$_{\textbf{2.13}}$} \\
& &F1 & 70.70 & 72.12 & 78.09&91.40 & \underline{92.10$_{\text{1.43}}$} & {70.14$_{{3.20}}$} & 92.05$_{\text{2.13}}$ & 91.14$_{\text{1.70}}$&  83.48$_{\text{3.96}}$ & \textbf{93.92$_{\textbf{1.73}}$} \\

\midrule
\multicolumn{2}{c}{\multirow{2}{*}{{Average}}}&Acc &70.58	&67.09	&76.83	&87.82&	\underline{89.36}&	71.70&	85.04&	88.89&	87.59&\textbf{92.21} \\

&&F1 &62.14&	64.20&	69.63&	87.06&	\underline{87.37}&	66.73&	82.94&	86.93&	85.29&	\textbf{91.96} \\

\bottomrule
\end{tabular}
}

\caption{{{Accruacy and F1 score (\%) of \modelname{} and baseline methods for MGT detection under the DG setting. }} 
The results are average values of 10 runs with different random seeds. The subscript means the standard deviation (\eg $93.92_{1.73}$  means 93.92 ± 1.73). 
Metric-based methods' results are deterministic, so we do not report standard deviation. Also, these metric-based methods must have the white-box generator as the base model, which is different from the model-based methods. 
$\dag$ denotes using the RoBERTa-base (125M) as the backbone model.
The best and second-best are \textbf{bolded} and \underline{underlined} respectively.
    }
\label{tab:ducomain}
\end{table*}

\section{Experiments} 

\subsection{Experiment Settings} 
To evaluate DG in MGT detection, we conduct extensive experiments on three open‑source datasets, including M4 \cite{wang2023m4}, DetectRL \cite{NEURIPS2024_b61bdf7e}, and MAGE \cite{li-etal-2024-mage} dataset, under three experimental settings:
\textbf{\textit{i}) Cross-Domain}, composed of the DetectRL and MAGE datasets, and including MGTs from Opinion Statement (Opinion.S), Question Answering (Question.A), Story generation (Story.G), and Scientific Writing (Scientific.W).
\textbf{\textit{ii) }Cross-Generator}, composed of the M4 and MAGE datasets, in which we obtain MGTs generated by Flan-T5 \cite{chung2024scaling}, ChatGPT \cite{OpenAI2022ChatGPT}, GLM \cite{zeng2022glm}, and LLaMA \cite{touvron2023llama}. 
\textbf{\textit{iii}) Cross-Scale}, based on the MAGE dataset, including MGTs generated by LLaMA2‑13B, LLaMA2‑30B, and LLaMA2‑65B.

We randomly select 1000 samples from the original training set, balanced across datasets, domains, and categories.  
Training is performed for 30 epochs using AdamW ($\epsilon=2\!\times\!10^{-5}$), 
with a learning rate of 0.01 and a scaling factor $\tau$ of 0.6.
More details are shown in the Appendix A. 
%

\subsection{Competitors} 

We evaluate \modelname{} against nine methods for MGT detection, including metric-based and fine-tuned detectors.

\fakeparagraph{Metric-based detectors} Log-probabilities from a generative LM are often used for classification with a predefined threshold \footnote{We utilizing the GPT-Neo-2.7B \cite{black2021gptneo} to align with Fast-DetectGPT \cite{bao2024fastdetectgpt} experiments.}, including \textit{GLTR} \cite{gehrmann2019gltr}, \textit{DetectGPT} \cite{mitchell2023detectgpt}, \textit{Fast-DetectGPT} \cite{bao2024fastdetectgpt}, \textit{Binoculars} \cite{hans2024spotting} \footnote{Under the original Falcon-7B and Falcon-7B-Instruct setting.}.

\fakeparagraph{Fine-tuned detectors} Supervised detectors trained on a PLM, typically optimized with a classification loss, including \textit{RoBERTa} \cite{liu2019roberta}, \textit{Ghostbuster} \cite{verma2023ghostbuster}, \textit{RADAR} \cite{hu2023radar}, \textit{PECOLA} \cite{liu2024does}, \textit{ImBD} \cite{chen2025imitate}.

\subsection{Performance Comparison}

We report the experimental results in Table~\ref{tab:ducomain}.
We first compare \modelname{} with state-of-the-art metric-based and model-based detectors.
In the DG setting, compared to the strongest baselines RoBERTa and Binoculars, \modelname{} achieves average improvements of 3.62\% in accuracy and 4.75\% in F1 score, demonstrating strong DG.
Broadly speaking, the mostly fine-tuned detectors outperform metric-based methods across all datasets because the performance significantly drops when the scoring model differs from the target model \cite{mitchell2023detectgpt}.
Moreover, the detection results in the cross-generator setting are generally lower than those in the cross-domain and cross-scale settings, likely due to more severe domain shifts introduced by different generators. 
It demonstrates that the different generation paradigms across models lead to a larger domain shift, degrading the model performance.
In the cross-domain setting, testing the model on the Scientific.W dataset leads to consistent low accuracy below 81\% because of the large difference between general writing and the scitific writing which uses more technical words \cite{liu2024does}.

\subsection{Ablation Study}
We conduct ablation experiments to understand the contribution of each component of \modelname in the cross-generator setting.
The core components we examine are:
Low Frequency Filtering Module (LFF), Frequency Spectrum Reconstruction (FSR), and Frequency Spectrum Alignment (FSA).
As shown in Table \ref{tab:blocksablation}, 
we find that every block in \modelname contributes to the DG of MGT detector, shown by the improvement over the  RoBERTa baseline.
Moreover, the FSA module contributes to the detection performance most, indicating the alignment of frequency-domain features plays an important role in mitigating domain shift.
In average, the combination of two components is always superior to only incorporating one module into the training process, indicating that different modules are complementary to each other and collectively boost the DG of MGT detector. We also discuss the effects of different scale factors and test lengths in Appendix B.1 and B.2. 

\begin{table}[t]
\centering
\renewcommand{\arraystretch}{1.2}
\resizebox{\columnwidth}{!}{
\begin{tabular}{ccc|cccc|c}
\toprule
\multicolumn{3}{c|}{\textbf{Modules}} & \multicolumn{5}{c}{\textbf{Cross-Generator}} \\
\hline
\textbf{LFF} & \textbf{FSR} & \textbf{FSA} & \textbf{Flan-T5} & \textbf{ChatGPT} & \textbf{GLM} & \textbf{LLaMa} &  \textbf{Avg.} \\
\midrule
-- & -- & -- & 85.95 & 84.65 & 86.95 & 80.85 & 84.60 \\
\textbf{\textcolor{green}{\checkmark}} & -- & -- & 87.84 & 85.61 & 88.61 & 84.92 & 86.75 \\
-- & \textbf{\textcolor{green}{\checkmark}} & -- & 86.40 & 84.82 & 89.15 & 83.91 & 86.07 \\
-- & -- & \textbf{\textcolor{green}{\checkmark}} & 88.05 & 85.27 & 89.01 & 85.28 & 86.90 \\
\midrule
\textbf{\textcolor{green}{\checkmark}} & \textbf{\textcolor{green}{\checkmark}} & -- & 87.42 & 86.49 & 90.02 & 85.81 & 87.44 \\
\textbf{\textcolor{green}{\checkmark}} & -- & \textbf{\textcolor{green}{\checkmark}} & 88.04 & 88.05 & 89.99 & 85.71 & 87.94 \\
-- & \textbf{\textcolor{green}{\checkmark}} & \textbf{\textcolor{green}{\checkmark}} & 87.21 & 86.85 & 89.25 & 86.93 & 87.56 \\
\textbf{\textcolor{green}{\checkmark}} & \textbf{\textcolor{green}{\checkmark}} & \textbf{\textcolor{green}{\checkmark}} & \textbf{89.02} & \textbf{89.23} & \textbf{92.82} & \textbf{87.20} & \textbf{89.57} \\
\bottomrule
\end{tabular}
}
\caption{\textbf{F1 score (\%) for Ablation study on different module combinations.}
\textcolor{green}{\checkmark} means that we keep the corresponding block and - means the block is removed.
}
\label{tab:blocksablation}
\vspace{-0.5cm}
\end{table}

\subsection{Discussion}
\fakeparagraph{Effect of Individual Frequency Components} 
Compared to style and theme, which are coarse-grained and unstable across domains or generators, sentence structure and token-level statistics provide more reliable signals for detection. While many existing methods exploit such fine-grained cues (e.g., LM probabilities and perturbations), our approach further aligns multi-granular features with frequency bands. 
As shown in Table \ref{tab:lowmidhigh}, results reveal that using only the low-frequency band leads to an average F1 score of 83.22\%, representing a 1.38\% drop compared to the all-features model. In contrast, isolating the mid- and high-frequency bands individually leads to performance gains of 1.01\% and 1.47\%, respectively, compared to the all-features model.

\begin{table}[ht]
\centering
\renewcommand{\arraystretch}{1.2}
\resizebox{\columnwidth}{!}{
\begin{tabular}{ccc|cccc|c}
\toprule
\multicolumn{3}{c|}{\textbf{Modules}} & \multicolumn{5}{c}{\textbf{Cross-Generator}} \\
\hline
\textbf{Low-} & \textbf{Mid-} & \textbf{High-} & \textbf{Flan-T5} & \textbf{ChatGPT} & \textbf{GLM} & \textbf{LLaMa} &  \textbf{Avg.} \\
\midrule

\textbf{\textcolor{green}{\checkmark}} & -- & -- & 85.01 & 83.64 & 85.29 & 79.96 & 83.22 \\
-- & \textbf{\textcolor{green}{\checkmark}} & -- & 86.01 & \textbf{85.31} & 88.64 & \textbf{83.49} & 85.61 \\
-- & -- & \textbf{\textcolor{green}{\checkmark}} &\textbf{ 87.65} & 85.07 &\textbf{ 89.49} & 83.08 &\textbf{ 86.07} \\
\midrule

\textbf{\textcolor{green}{\checkmark}} & \textbf{\textcolor{green}{\checkmark}} & \textbf{\textcolor{green}{\checkmark}} & {85.95} & {84.65} & {86.95} & {80.85} & {84.60} \\
\bottomrule
\end{tabular}
}
\caption{F1 score from Ablation on Individual Components.
}
\label{tab:lowmidhigh}
\end{table}

\fakeparagraph{Robustness under Perturbation}
We test the robustness of detectors under four attacks (\ie Delete, Insert, Repeat, and Generate) with a perturbation rate of 15\%, following \citet{wang-etal-2024-stumbling}.
As shown in Table~\ref{tab:perturbation}, our method exhibits a consistently smaller drop in F1 score than the baselines RoBERTa and Binoculars. \modelname{} achieves an average improvement of 5.29\% in F1 score.
Specifically, under the deletion and repetition perturbations, the averaged F1 score of \modelname{} decreases by only 5.10\% and 5.23\% in the two scenarios (\ie cross-generator and cross-domain), underscoring its remarkable robustness. 
The complete evaluation results for all attacks are provided in the Appendix C.1.

\begin{table}[ht]
\centering
\resizebox{\columnwidth}{!}{
\begin{tabular}{l|cccc|c}
\toprule
\textbf{Perturbation} & \textbf{Delete} & \textbf{Insert} & \textbf{Repeat} & \textbf{Generate} & \textbf{Avg.} \\
\midrule

\multicolumn{6}{c}{\textbf{\textit{Binoculars}}} \\
\cmidrule{1-6}
Cross-Generator & 65.41& 65.12 & 65.66 & 59.59 & 63.95 \\
Cross-Domain    & 83.97 & 81.10 & 82.45 & 67.45 & \underline{78.74}\\
Cross-Scale    & 84.78 & 76.19 & 68.53 & 61.55 & \underline{72.76} \\

\midrule
\multicolumn{6}{c}{\textbf{\textit{RoBERTa-base}}} \\
\cmidrule{1-6}
Cross-Generator & 75.065 & 59.91 & 81.285 & 60.03 & \underline{69.07}\\
Cross-Domain    & 85.35 & 74.81 & 81.40 & 72.90 & {78.61} \\
Cross-Scale    & 83.15 & 73.22 & 68.16 & 65.06 & {72.39} \\

\midrule
\multicolumn{6}{c}{\textbf{\textit{\modelname{}}}} \\
\cmidrule{1-6}
Cross-Generator & 79.61 & 67.99 & 84.58 & 67.40 & \textbf{74.90} \\
Cross-Domain    & 89.32 & 77.12 &85.12 & 77.67 & \textbf{82.30 }\\
Cross-Scale    & 85.74 & 76.53 & 73.95 & 69.51 & \textbf{76.43} \\

\bottomrule
\end{tabular}
}
\caption{\textbf{Performance on diverse perturbation attacks.}
Results are reported as the average F1 score (\%) across three generalization settings and four perturbation scenarios. GPT-2 XL (1.5B) \cite{solaiman2019release} is employed to construct perturbations, including  generating and inserting. 
}
\label{tab:perturbation}
\end{table}

\fakeparagraph{Applicability to Different Backbones} 
We evaluate the effectiveness of \modelname{} with RoBERTa-large and BERT-large and Qwen3-0.6B \cite{qwen3embedding} as the backbones in both in-domain (IND) and  DG settings.
As shown in \tabref{tab:OOD}, \modelname always outperforms vanilla fine-tuning with different backbones, in all settings.
Specifically, \modelname improves over the vanilla fine-tuning by 1.16\% and 2.11\% in RoBERTa-large, and by an average of 1.53\% on Qwen3-0.6B.
The results demonstrate that \modelname is applicable to different backbones with different scales.
More test results are shown in Appendix  C.2.
We also visualization of feature distribution in Appendix~C.3.


\begin{table}[ht]
\centering
\resizebox{\columnwidth}{!}{
\begin{tabular}{ll|ccc|c}
\toprule
\multicolumn{6}{c}{\textbf{\textit{In-domain (IND)}}} \\
\midrule
\textbf{Model} & \textbf{Method} & \textbf{Generator} & \textbf{Domain} & \textbf{Scale} & \textbf{Avg.} \\
\midrule
\multirow{2}{*}{RoBERTa-base} 
& base  & 93.45 & 96.01 & 94.90 & 94.78 \\
& \modelname{} &{94.89} & {97.43} & {95.32} & {95.88} \\
\midrule
\multirow{2}{*}{RoBERTa-large} 
& base  & 94.75 & 97.87& 96.24 & 96.28 \\
& \modelname{} &\textbf{96.05} & \textbf{98.98} & \textbf{97.29} & \textbf{97.44} \\
\midrule
\multirow{2}{*}{BERT-large} 
& base  &88.21  &94.49 &92.23  &91.64  \\
& \modelname{} &90.01 &95.24  &92.94  &92.73  \\
\midrule
\multirow{2}{*}{QWen3-0.6B} 
& base  &93.81  &96.82 &95.82  &95.48  \\
& \modelname{} &\underline{95.28} &\underline{98.24}  &\underline{96.49}  &\underline{96.67} \\

\midrule
\multicolumn{6}{c}{\textbf{\textit{Domain Generalization (DG)} }} \\
\midrule
\multirow{2}{*}{RoBERTa-large} 
& base  & 88.13 & \underline{93.10} & 93.81 & {91.49}\\
& \modelname{} & \textbf{90.84} &\textbf{95.02} & \textbf{95.37} & \textbf{93.60}\\
\midrule
\multirow{2}{*}{BERT-large} 
& base  &81.4  &82.68 & 89.88 & 84.17 \\
& \modelname{} &84.18 &88.71  & 92.97 & 88.22 \\
\midrule
\multirow{2}{*}{QWen3-0.6B} 
& base  &86.17  &89.08 &93.86  &89.32 \\
& \modelname{} &\underline{87.47} &92.04  & \underline{95.03}  &\underline{91.19}  \\
\bottomrule
\end{tabular}
}
\caption{\textbf{F1 score (\%) Comparison with Different Backbones.} For the DG setting, the reported results are averaged across all test subsets. For the IND setting, all subcategories are merged for training and testing.
}
\label{tab:OOD}
\end{table}

\begin{figure}[H]
  \centering
  \includegraphics[width=0.9\columnwidth]{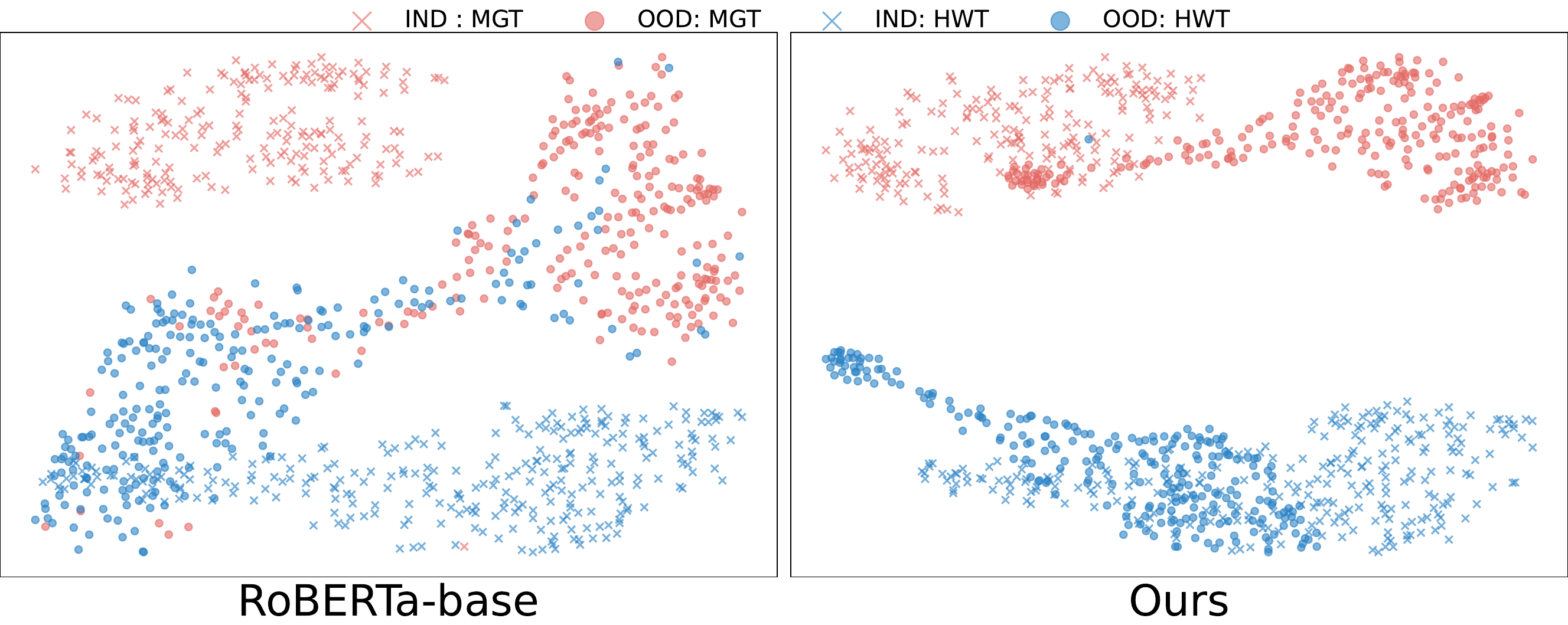}
  \caption{  \textbf{t-SNE Visualization of the Learned [CLS] Embeddings.}
    Comparison between RoBERTa-base and \modelname{} on IND and DG test sets. 
}
  \label{fig:gpt} 
\end{figure}

\fakeparagraph{Visualization of Feature Distribution}  
To better understand the improved generalization performance of our method in detecting MGT under domain shifts, we extract features of the test samples from the final hidden layer of both RoBERTa-base and \modelname{}, and use t-SNE~\cite{van2008visualizing} to project them into a 2D space for visualization, as illustrated in \figref{fig:gpt}.
Under the DG setting, we use the ChatGPT subset as test data and observe that \modelname{} produces representations with clearer separation between HWT and MGT instances than RoBERTa-base. Moreover, data points with the same label are more tightly clustered.
More visualizations across different datasets are reported in Appendix C.3.

\section{Conclusion}
In this work, we focus on enhancing the domain generalization ability of MGT detectors with concentration on mitigating domain shift brought by data sources.
We transfer the text representation to frequency domain using DFT and find that the spectra of text features mitigate the domain shift among intra-class samples.
Based on the decomposability of frequency domain, we design a low-frequency spectrum filtering module to remove the low-frequency feature which is more affected by the domain shift.
Moreover, we align the spectra of intra-class samples from different domains with frequency spectrum alignment module to mitigate the domain shift, therefore enabling the model to learn task-specific and domain-invarient features from frequency domain.
Experimental results show that our method outperforms SOTA metric-based and model-based baselines.
We hope that our work can inspire future research in AI-generated content detection in other modalities, 
and serve as a foundation for developing a unified detecting approach.

\section*{Ethics Statement}

\modelname{} is designed to assist users in reasonably and accurately identifying MGT. Our goal is to develop a generalizable method that can be extended to other modalities, such as image and audio, and to inspire progress toward more robust MGT detectors. While we aim to support the responsible development and deployment of detection techniques, we explicitly discourage any malicious use of our work, particularly attempts to evade or undermine detection systems. All datasets used in this study are publicly available.

\newpage
\bibliography{custom}
\bibliographystyle{acl_natbib}

\appendix

\clearpage

\section{Implementation Details}
\label{app:implementation}
This section presents the hyperparameter settings, meta-information, and the learning process of our framework for the DG task in MGT detection.
All the experiments are conducted on a single NVIDIA A100 80G GPU.

\subsection{Hyperparameter Details}
\label{app:details}
In this subsection, we provide additional details on hyperparameter settings to support the reproducibility of our method. For each experiment, we randomly select 10 different seeds and report the average test accuracy, as shown in Table~\ref{tab:Detail}.

\begin{table}[H]
\centering
\tabcolsep=3pt
{\fontsize{11pt}{12pt}\selectfont
\begin{tabular}{lc} 
\toprule
\textbf{Parameter} & \textbf{Value} \\
\midrule
Training Epochs & 30 \\
Optimizer & AdamW \\ 
Learning rate & 2e-5 \\
Weight Decay & 0.01 \\
Batch Size & 32 \\ 
Pre-trained model  \ & \ RoBERTa-base \\ 
Scaling Factor $\tau$\ & 0.6 \\

\bottomrule
\end{tabular}
}
\caption{Implementation details of hyperparameters.}
\label{tab:Detail}
\end{table}

\subsection{Dataset Details}
We evaluate \modelname{} under three generalization settings (\ie Cross-Generator, Cross-Domain, Cross-Scale) and 11 test sets in total, as shown in Table~\ref{tab:mega_data}.
\label{app:mega_data}

\begin{table*}[ht]
\centering
\resizebox{\textwidth}{!}{
\begin{tabular}{cc|cccc|cccc|ccc}
\toprule
\multicolumn{2}{c|}{\textbf{Dataset}} 
& \multicolumn{4}{c|}{\textbf{Cross-Generator}} 
& \multicolumn{4}{c|}{\textbf{Cross-Domain}} 
& \multicolumn{3}{c}{\textbf{Cross-Scale}}  \\

\midrule
&{{Split}}
& {{\textit{FLAN-T5}}} 
& {{\textit{ChatGPT}}} 
& {{\textit{GLM}}} 
& {{\textit{LLaMA}}} 
& {{\textit{Opinion.S}}} 
& {{\textit{Question.A}}} 
& {{\textit{Scientific.W}}} 
& {{\textit{Story.G}}} 
& {{\textit{LLaMa-13B}}} 
& {{\textit{LLaMa-30B}}} 
& {{\textit{LLaMa-65B}}} 
\\
\toprule
&Train  
& 1000   &1000    &1000   &1000  
&1000   & 1000 &  1000  &   1000
&  1000  &  1000  & 1000
\\
&Valid
&  3000  &  3000  &  3000 &  3000
&  2000 & 2000 &  2000  &   2000
&   3000 &  3000  & 3000
\\
&Test
&  6000  &  6000  & 6000  &  6000
&  4000 &4000  &  4000  &   4000
&  6000  &  6000  & 6000
\\
\bottomrule
\end{tabular}
}
\caption{Number of instances across three generalization settings and eleven test sets.
    }
\label{tab:mega_data}

\end{table*}

\subsection{Learning Process}
Our MGT detection framework integrates a Frequency Feature
Alignment Loss ($\mathcal{L}_{\text{MAE}}$) and a feature space loss ($\mathcal{L}_{\text{CE}}$), as detailed in Algorithm~\ref{algorithm:aligned}.
It also demonstrates that our method can be extended to a variety of classification tasks to enhance domain generalization.
\label{app:Process}

\begin{algorithm}[t]
\caption{Learning Process}
\label{algorithm:aligned}
\begin{algorithmic}[1]
\State \textbf{Input:}  training dataset $\mathcal{D}_{train}$,Valid dataset $D_{\text{val}}$, learning rate $lr$, Scaling Factor $\tau$, Detector model
\State Initialize the pretrained Transformer $M$
\State Compute the global frequency-spectrum distribution over the training set $\mathcal{D}_{\text{train}}$.
\For{epoch $e = 1...epoch$}
    \For{batch $b = 1...Batch$}
        \State Get Filtered Frequency $\mathbf{H}_{\text{f}}$ by Eq. (9)
        \State  Reconstruct $\mathbf{H}_{\text{f}}$ by Eq. (12)
    \EndFor
    \State Inverse Fourier transform by Eq. (13)
    \State Calculate frequency loss $\mathcal{L}_{\mathrm{MAE}}$ by Eq. (16)
    \State Calculate the classification loss $\mathcal{L}_{CE}$ on the reconstructed features
    \State Calculate $\mathcal{L} = \mathcal{L}_{CE} + \mathcal{L}_{\mathrm{MAE}}$
    \State Update detector model parameters w.r.t.$\mathcal{L}$ via backpropagation
    \If{$e$ \% Eval steps=0}  
    Fit and Evaluate the DG detector model on $D_{\text{val}}$
    \EndIf
    \State  Return the best detector model checkpoint.
\EndFor
\State \textbf{Output:} A trained and aligned DG detector model.
\end{algorithmic}
\end{algorithm}

\subsection{Competitors} 

We evaluate \modelname{} against nine methods for MGT detection, including metric-based and fine-tuned detectors.

\textit{GLTR} \cite{gehrmann2019gltr}, a statistical tool that visualizes token-level prediction probabilities to detect AI-generated content.

\textit{DetectGPT} \cite{mitchell2023detectgpt}, a zero-shot metric-based detector that leverages the curvature of the log probability function to differentiate in MGT detection.

\textit{Fast-DetectGPT} \cite{bao2024fastdetectgpt}, an optimized version of DetectGPT, 
it offering significant speedup  while maintaining or surpassing the detection accuracy of DetectGPT.

\textit{Binoculars} \cite{hans2024spotting}, computes perplexity and cross-perplexity with two models for MGT detection. \footnote{Following the original setting, we use  Falcon-7B and Falcon-7B-Instruct \cite{almazrouei2023falcon} respectively.}

\textit{RoBERTa} \cite{liu2019roberta}, a transformer-based model fine-tuned for binary classification.


\textit{Ghostbuster} \cite{verma2023ghostbuster}, a feature-based approach that processes text with multiple weaker models.

\textit{RADAR} \cite{hu2023radar}, based adversarial learning to train a paraphraser and a detector simultaneously.

\textit{PECOLA} \cite{liu2024does}, a contrastive learning-based model that enhances perturbation-based detection by introducing selective perturbation method.

\textit{ImBD} \cite{chen2025imitate}, a threshold-based model that leverages style-conditioned probability curvature for detection aligned with LLMs. 
\footnote{Following the original setting \cite{chen2025imitate}, we fine-tune the GPT-Neo-2.7B \cite{black2021gptneo} scoring model on the training set with style preference optimization. The optimal threshold is determined by maximizing accuracy on the training set and applied unchanged during test set evaluation.}

\section{Efficiency of \modelname{}}
\label{app:Efficiency}
\subsection{Effect of Scaling Factor}
The scaling factor \( \tau \) is used to construct a multi-scale frequency-domain structure.
A lower value of $\tau$ corresponds to a broader mid-frequency range.
We evaluate \( \tau \) in the cross-generator setting, and the average accuracy results in \figref{fig:scalingfactor}(a) show that, as \( \tau \) increases from 0.1 to 1.0, the F1 score remains within the range of 87.60\% to 89.56\%, consistently outperforming the baseline.
This demonstrates the stability and effectiveness of our method. The best performance is achieved when \( \tau = 0.6 \), indicating that introducing a scaling factor can significantly enhance model generalization. Accordingly, we set \( \tau = 0.6 \) in all our experiments. 

\subsection{Effect of Text Length}

\begin{figure}[h]
  \centering
  \includegraphics[width=\columnwidth]{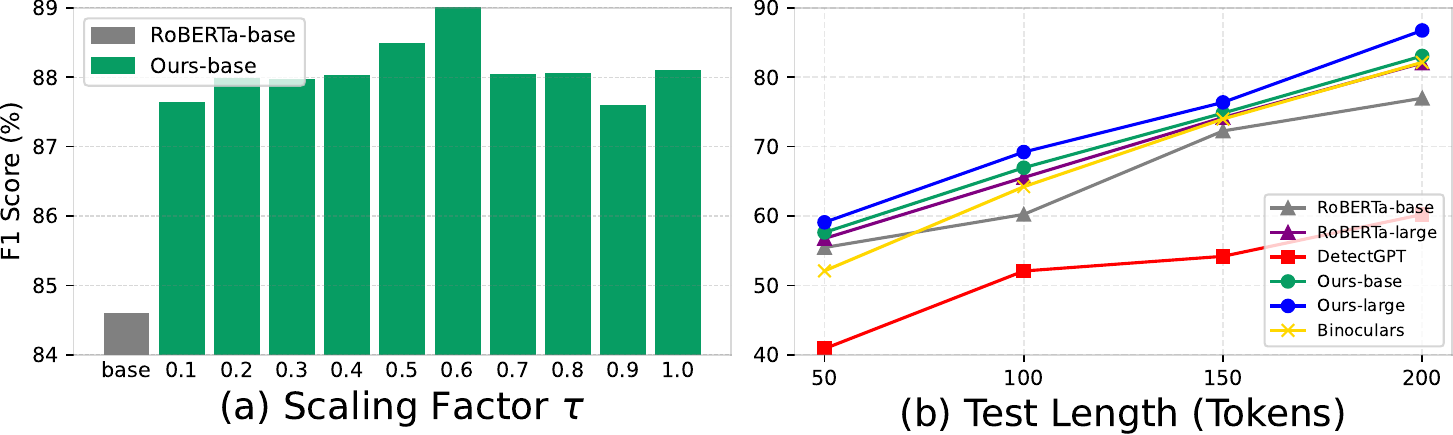}
  \caption{\textbf{Effects of hyperparameter $\tau$ and test length.} Results are averaged over the cross-generator setting, using RoBERTa-base and RoBERTa-large as backbones and denoted as Ours-base and Ours-large, respectively. 
  }
  \label{fig:scalingfactor} 

\end{figure}

To examine the ability of \modelname{} to detect short MGTs, we chunk the test samples into segments of 50, 100, 150, and 200 tokens under the cross-generator setting.
As shown in \figref{fig:scalingfactor}(b), \modelname{} consistently outperforms SOTA methods across all text lengths, with accuracy showing a steady improvement as the input length increases.

\subsection{Analysis of Domain Shift}
 MAE is a common metric for measuring the deviation between predicted and true values \cite{yi2023frequency,yi2024filternet}. 
We use MAE to quantify the domain shift of intra-class data to reveal the reason why our method performs well in domain generalization.
As shown in \tabref{tab:mae_cross}, in cross-generator, our method \modelname{} achieves the lowest MAE between training and testing samples with the same label, demonstrating effective reduction of domain shift for MGT in the DG detection process.
Moreover, the average MAE between different-label pairs under \modelname{} is 0.0395 higher than that of RoBERTa-base in test data, highlighting better feature separability across different types of text. 
Extended MAE results for the remaining two DG scenarios (\ie cross-domain, cross-scale) are provided in \tabref{tab:mae_cross_all}.

\begin{table}[t]
\centering
\renewcommand{\arraystretch}{1.2}
\resizebox{\columnwidth}{!}{
\begin{tabular}{ll|cccc|c}
\toprule
\textbf{Method} & \textbf{Dataset} & \textbf{Flan-T5} & \textbf{ChatGPT} & \textbf{GLM} & \textbf{LLaMA} & \textbf{Avg.} \\
\midrule
\multirow{4}{*}{\textit{RoBERTa-base}} 
  & $\text{Train}_m\!:\!\text{Test}_m$ & 0.1521 & 0.1296 & 0.1424 & 0.1364 & 0.1401 \\
  & $\text{Train}_m\!:\!\text{Test}_h$ & 0.5071 & 0.4720 & 0.4612 & 0.4379 & \underline{0.4696} \\
  & $\text{Test}_m\!:\!\text{Test}_h$  & 0.2881 & 0.3044 & 0.3527 & 0.3181 & 0.3158 \\
  & $\text{Train}_h\!:\!\text{Test}_h$ & 0.1127 & 0.1023 & 0.1412 & 0.1322 & 0.1221 \\
\midrule
\multirow{4}{*}{\textit{\modelname{}}} 
  & $\text{Train}_m\!:\!\text{Test}_m$ & 0.1092 & 0.1132 & 0.0985 & 0.1299 & \textbf{0.1127} \\
  & $\text{Train}_m\!:\!\text{Test}_h$ & 0.4239 & 0.4374 & 0.4497 & 0.4787 & 0.4474 \\
  & $\text{Test}_m\!:\!\text{Test}_h$  & 0.4223 & 0.3345 & 0.3288 & 0.3356 & \underline{0.3553} \\
  & $\text{Train}_h\!:\!\text{Test}_h$ & 0.0628 & 0.0902 & 0.0693 & 0.0997 &\textbf{0.0805} \\
\bottomrule
\end{tabular}
}
\caption{\textbf{Comparison of MAE distributions.}
\(\text{Train}_m\) and \(\text{Test}_h\) denote the training and testing set composed of MGT and HWT data, respectively. A lower MAE indicates better alignment between training and testing distributions for the same label. The minimum MAE among same-label pairs in \textbf{bold}, while the maximum MAE among different-label pairs in \underline{underline}.
}
\label{tab:mae_cross}
\end{table}

\begin{table*}[t]
\centering
\renewcommand{\arraystretch}{1.2}
\resizebox{\textwidth}{!}{
\begin{tabular}{l|l|cccc|c|ccc|c}
\toprule
\multicolumn{2}{c|}{\textbf{Dataset}} 
& \multicolumn{5}{c|}{\textbf{Cross-Domain}} 
& \multicolumn{4}{c}{\textbf{Cross-Scales}} \\
\cmidrule(lr){1-2} \cmidrule(lr){3-7} \cmidrule(lr){8-11}
\textbf{Model} & \textbf{Split} 
& \textbf{Opinion.S} & \textbf{Question.A} & \textbf{Scientific.W} & \textbf{Story.G} & \textbf{Avg.} 
& \textbf{LLaMA-13b} & \textbf{LLaMA-30b} & \textbf{LLaMA-65b} & \textbf{Avg.} \\
\midrule
\multirow{4}{*}{\textit{RoBERTa-base}} 
  & $\text{Train}_m\!:\!\text{Test}_m$ & 0.0772 & 0.1359 & 0.3565 & 0.0983 & 0.1670 & 0.0632 & 0.0569 & 0.0792 & 0.0664 \\
  & $\text{Train}_m\!:\!\text{Test}_h$ & 0.4484 & 0.4321 & 0.3851 & 0.3649 & \textcolor{blue}{0.4076} & 0.4775 & 0.4573 & 0.4624 & 0.4657 \\
  & $\text{Test}_m\!:\!\text{Test}_h$  & 0.4242 & 0.3912 & 0.1742 & 0.4036 & 0.3483 & 0.4377 & 0.4014 & 0.3929 & 0.4107 \\
  & $\text{Train}_h\!:\!\text{Test}_h$ & 0.0890 & 0.0879 & 0.3372 & 0.1103 & 0.1561 & 0.0278 & 0.0893 & 0.0875 & 0.0682 \\
\midrule
\multirow{4}{*}{\textit{\modelname{}}} 
  & $\text{Train}_m\!:\!\text{Test}_m$ & 0.0422 & 0.0662 & 0.2643 & 0.0954 & \textcolor{red}{0.1170} & 0.0301 & 0.0566 & 0.0480 & \textcolor{red}{0.0449} \\
  & $\text{Train}_m\!:\!\text{Test}_h$ & 0.4607 & 0.4701 & 0.2432 & 0.4072 & 0.3953 & 0.4538 & 0.4627 & 0.4622 & \textcolor{blue}{0.4596} \\
  & $\text{Test}_m\!:\!\text{Test}_h$  & 0.4402 & 0.4244 & 0.1729 & 0.4335 & \textcolor{blue}{0.3678} & 0.4405 & 0.3923 & 0.4336 & \textcolor{blue}{0.4221} \\
  & $\text{Train}_h\!:\!\text{Test}_h$ & 0.0622 & 0.0334 & 0.1324 & 0.0789 & \textcolor{red}{0.0767} & 0.0075 & 0.0723 & 0.0371 & \textcolor{red}{0.0390} \\
\bottomrule
\end{tabular}
}
\caption{\textbf{Comparison of MAE distributions.} We compute the MAE under both Cross-Domain and Cross-Scale settings. $\text{Train}_h$ and $\text{Test}_m$ denote training and testing sets composed of HWT and MGT, respectively. The minimum MAE among same-label pairs in \textcolor{red}{red}, while the maximum MAE among different-label pairs in \textcolor{blue}{blue}.}
\label{tab:mae_cross_all}

\end{table*}

\subsection{Frequency Component Construction under DG}
\label{app:Component}
As shown in Table~\ref{tab:multiscale}, we present multi-scales frequency estimates across three generalization scenarios. Sentence- and token-level results are averaged, and a scaling factor $\tau$ is introduced to adaptively align frequency boundaries with semantic structure levels (\eg paragraph-, sentence-, and token-level granularity) across different samples. Low-frequency components concentrate around indices 0–2, capturing slowly varying signals over long input spans.

\begin{table*}[ht]
\centering
\renewcommand\arraystretch{1.25}
\resizebox{\textwidth}{!}{
\begin{tabular}{c|ccc|ccc}
\toprule
\multirow{2}{*}{\textbf{Dataset}} & \multicolumn{3}{c|}{\textbf{MGT}} & \multicolumn{3}{c}{\textbf{HWT}} \\
\cmidrule(lr){2-4} \cmidrule(lr){5-7}
& Frequency-Domain & Multi-Scale (Average) & FFT Bin 
& Frequency-Domain & Multi-Scale (Average) & FFT Bin \\
\midrule

\multirow{3}{*}{\textbf{Cross-generator}}
& Low-  & Paragraph     & 0--1     & Low-  & Paragraph     & 0--1     \\
& Mid-  & Sentence: 28  & 2--162   & Mid-  & Sentence: 26  & 2--162   \\
& High- & Tokens: 412   & 162--384 & High- & Tokens: 401   & 163--384 \\
\midrule

\multirow{3}{*}{\textbf{Cross-Scales}}
& Low-  & Paragraph     & 0--2     & Low-  & Paragraph     & 0--2     \\
& Mid-  & Sentence: 20  & 3--164   & Mid-  & Sentence: 24  & 3--162   \\
& High- & Tokens: 294   & 1165--384 & High- & Tokens: 323   & 163--384 \\
\midrule

\multirow{3}{*}{\textbf{Cross-Domain}}
& Low-  & Paragraph     & 0--1     & Low-  & Paragraph     & 0--1     \\
& Mid-  & Sentence: 30  & 2--161   & Mid-  & Sentence: 25  & 2--161   \\
& High- & Tokens: 443   & 163--384 & High- & Tokens: 396   & 162--384 \\
\bottomrule
\end{tabular}
}
\caption{Multi-scale frequency estimation based on token and sentence statistics under different generalization settings for MGT and HWT ($\tau=0.6$). Due to the Hermitian symmetry of the frequency spectrum for real-valued signals, we visualize only the first $\left\lfloor \frac{N}{2} \right\rfloor + 1$ frequency bins, which sufficiently represent the full spectrum.
}. 
\label{tab:multiscale}

\end{table*}

\section{Additional Experiment Results}

\subsection{Robustness under Perturbation}

Table~\ref{tab:all_pertubation} shows the performance drop for four individual typo perturbation types, \ie insert, delete, repeat, and generate.  In both the cross-domain and cross-scale settings, these perturbations exhibit similar degradation trends, consistent with the results observed in the cross-generator experiments.

\label{app:Robustness_complete}

\begin{table*}[ht]
\centering
\setlength{\tabcolsep}{2pt} 
\renewcommand\arraystretch{1.2}
\resizebox{\textwidth}{!}{
\begin{tabular}{cc|ccccc|ccccc|ccccc}
\toprule
\multicolumn{2}{c|}{\textbf{Cross-}} 
& \multicolumn{5}{c|}{\textbf{Binoculars}} 
& \multicolumn{5}{c|}{\textbf{RoBERTa}} 
& \multicolumn{5}{c}{\textbf{\modelname{}}} \\

\midrule
\textbf{Dataset} 
&{\textbf{Test data}}
& {\textbf{\textit{Delete}}} 
& {\textbf{\textit{Insert}}} 
& {\textbf{\textit{Repeat}}} 
& {\textbf{\textit{Generate}}} 
& {\textbf{\textit{Avg.}}}
& {\textbf{\textit{Delete}}} 
& {\textbf{\textit{Insert}}} 
& {\textbf{\textit{Repeat}}} 
& {\textbf{\textit{Generate}}} 
& {\textbf{\textit{Avg.}}}
& {\textbf{\textit{Delete}}} 
& {\textbf{\textit{Insert}}} 
& {\textbf{\textit{Repeat}}} 
& {\textbf{\textit{Generate}}}
& {\textbf{\textit{Avg.}}}
\\
\toprule
\multirow{4}{*}{\textbf{\rotatebox{90}{Generator}}}
&{\textbf{FLAN-T5}}   
& 41.52 & 63.43 & 41.55 & 60.25 & 51.69
& 78.29 & 66.43 & 83.45 & 69.21  & \underline{74.34}
& 86.43 & 80.43 & 86.37 & 84.22  & \textbf{84.36}
\\

&{\textbf{ChatGPT}} 
& 71.82 & 63.30 & 69.70 & 57.20 & 65.5
& 75.51 & 56.28 & 78.87 & 54.65 & \underline{66.32}
& 77.45 & 62.50 & 83.46 & 60.56 & \textbf{70.99}
\\

&{\textbf{GLM}}     
& 79.12 & 70.22 & 69.23 & 60.15 & \underline{69.68}
& 76.34 & 56.25 & 82.34 & 56.15 & 67.77
& 80.15 & 65.76 & 84.32 & 60.10 & \textbf{72.58}
\\

&{\textbf{LLaMA}}     
& 69.21 & 63.56 & 82.15 & 60.75 & \underline{68.92}
& 70.12 & 60.68 & 80.48 & 60.10 & 67.85
& 74.43 & 63.28 & 84.18 & 64.71 & \textbf{71.65}
\\

\midrule
\multirow{4}{*}{\textbf{\rotatebox{90}{Domain}}}
&{\textbf{Opinion.S}}    
& 85.01 & 80.22 & 83.20 & 71.09 & \underline{79.88}
& 90.20 & 73.61 & 80.58 & 71.11 & 78.88
& 92.81 & 77.87 & 84.27 & 79.62 & \textbf{83.64}
\\
&{\textbf{Question.A}}      
& 90.13 & 88.12 & 89.57 & 66.71 & \underline{83.63}
& 90.41 & 79.45 & 90.36 & 72.92 & 83.29
& 93.94 & 78.89 & 92.82 & 77.28 & \textbf{85.73}
\\
&{\textbf{Scientific.W}} 
& 72.29 & 65.79 & 70.71 & 61.30 & 67.52
& 70.67 & 63.72 & 67.92 & 69.12 & \underline{67.85}
& 77.53 & 66.27 & 74.64 & 73.30 & \textbf{72.93}
\\
&{\textbf{Story.G}}     
& 88.45 & 90.30 & 86.32 & 70.70  & 83.94
& 90.14 & 82.44 & 86.72 & 78.46  & \underline{83.94}
& 92.99 & 85.46 & 88.75 & 80.49  & \textbf{86.92}
\\

\midrule
\multirow{3}{*}{\textbf{\rotatebox{90}{scale}}}
&{\textbf{LLaMa-13b}} 
& 86.75 & 80.01 & 70.21 & 57.49  & 73.61
& 87.15 & 71.25 & 72.32 & 65.28  & \underline{74.00}
& 88.48 & 72.85 & 77.36 & 70.29  & \textbf{77.24}
\\
&{\textbf{LLaMa-30b}} 
& 87.30 & 71.20 & 72.30 & 62.09  & \underline{73.22}
& 82.37 & 70.11 & 63.88 & 63.00  & 69.84
& 88.26 & 73.48 & 74.23 & 68.96  & \textbf{76.23}
\\
&{\textbf{LLaMa-65b}} 
& 80.29 & 77.35 & 63.09 & 65.07  & 71.45
& 79.93 & 78.29 & 68.29 & 66.89  & \underline{73.35}
& 80.47 & 83.26 & 70.27 & 69.28  & \textbf{75.82}
\\
\bottomrule
\end{tabular}
}

\caption{Detectors’ performance drops in terms of relative F1 score (\%) under four types of typo perturbations, namely insert, delete, repeat, and generate. The best and second-best are \textbf{bolded} and \underline{underlined} respectively.
    }
\label{tab:all_pertubation}

\end{table*}

\subsection{Additional Results for Applicability in Larger Models}
\label{app:Applicability_complete}
As shown in Table~\ref{tab:all_larger}, \modelname{} outperforms the base model across all 11 test settings when using either RoBERTa-large or BERT-large as the backbone. However, fine-tuned detectors based on BERT-large generally underperform those based on RoBERTa-large.
\begin{table*}[ht]
\centering
\setlength{\tabcolsep}{2pt} 
\renewcommand\arraystretch{1.2}
\resizebox{\textwidth}{!}{
\begin{tabular}{ccc|cccc|cccc|ccc}
\toprule
\multicolumn{3}{c|}{\textbf{Dataset}} 
& \multicolumn{4}{c|}{\textbf{\textbf{Cross-Generator}}} 
& \multicolumn{4}{c|}{\textbf{\textbf{Cross-Domain}}} 
& \multicolumn{3}{c}{\textbf{Cross-Scale}} \\
\midrule
\textbf{Model} 
&{\textbf{Method}}
&{\textbf{Metric}}
& {\textbf{\textit{FLAN-T5}}} 
& {\textbf{\textit{ChatGPT}}} 
& {\textbf{\textit{GLM}}} 
& {\textbf{\textit{LLaMA}}} 
& {\textbf{\textit{Opinion.S}}} 
& {\textbf{\textit{Question.A}}} 
& {\textbf{\textit{Scientific.W}}} 
& {\textbf{\textit{Story.G}}} 
& {\textbf{\textit{LLaMa-13B}}} 
& {\textbf{\textit{LLaMa-30B}}} 
& {\textbf{\textit{LLaMa-65B}}} \\
\toprule

\multirow{4}{*}{\textbf{{RoBERTa-large}}}
&\multirow{2}{*}{{\textbf{Base}}}
&Acc
& 88.21 & 88.54 & 89.26 & 85.98
& 95.60 & 95.95 & 80.26 & 98.02
& 94.11 & 93.28 & 92.92 \\
&&F1
& 87.99 & 88.75 & 90.16 & 85.64
& 95.70 & 95.75 & 82.06 & 98.92
& 94.61 & 93.80 & 93.02 \\

&\multirow{2}{*}{{\textbf{\modelname{}}}}
&Acc
& 91.76 & 90.44 & 92.26 & 89.71
& 97.51 & 97.55 & 88.70 & 99.16
& 96.82 & 95.08 & 95.01 \\
& &F1
& 91.06 & 90.55 & 92.06 & 89.70
& 97.90 & 97.25 & 85.70 & 99.26
& 96.72 & 94.78 & 94.61 \\

\midrule

\multirow{4}{*}{\textbf{{BERT-large}}}
&\multirow{2}{*}{{\textbf{Base}}}
&Acc
& 82.65 & 81.62 & 85.01 & 77.21
& 88.91 & 90.03 & 62.10 & 92.14
& 90.04 & 88.96 & 90.05 \\
& &F1
& 82.05 & 82.15 & 84.75 & 76.65
& 88.74 & 90.13 & 59.74 & 92.11
& 90.24 & 89.66 & 89.75 \\

&\multirow{2}{*}{{\textbf{\modelname{}}}}
&Acc
& 84.36 & 84.48 & 87.25 & 83.04
& 92.04 & 92.89 & 70.02 & 96.24
& 93.95 & 92.62 & 92.75 \\
&&F1
& 84.05 & 83.80 & 86.75 & 82.14
& 92.44 & 92.76 & 72.72 & 96.94
& 93.85 & 92.71 & 92.35 \\

\midrule

\multirow{4}{*}{\textbf{{QWen3-0.6B}}}
&\multirow{2}{*}{{\textbf{Base}}}
&Acc
& 85.88 & 86.43 & 90.02 & 83.45
& 93.60 & 95.95 & 79.66 & 97.26
& 95.02 & 93.95 & 93.46 \\
&&F1
& 86.06 & 86.25 & 89.12 & 83.25
& 93.00 & 95.65 & 70.61 & 97.06
& 94.88 & 93.65 & 93.06 \\

&\multirow{2}{*}{{\textbf{\modelname{}}}}
&Acc
& 87.92 & 87.88 & 90.46 & 85.01
& 95.33 & 96.46 & 81.26 & 98.45
& 95.71 & 94.21 & 94.05 \\
&&F1
& 87.20 & 87.95 & 90.16 & 84.60
& 95.10 & 96.16 & 78.16 & 98.75
& 95.45 & 94.91 & 94.75 \\

\bottomrule
\end{tabular}
}

\caption{Detection performance across out-of-domain (OOD) settings on three key evaluation scenarios and 11 subsets. The best are \textbf{bolded}.}
\label{tab:all_larger}
\end{table*}

\subsection{Additional Results for Visualization of Feature Distributions}
\label{app:Distribution_complete}
\begin{figure*}[t]
    \centering
    \includegraphics[width=\textwidth]{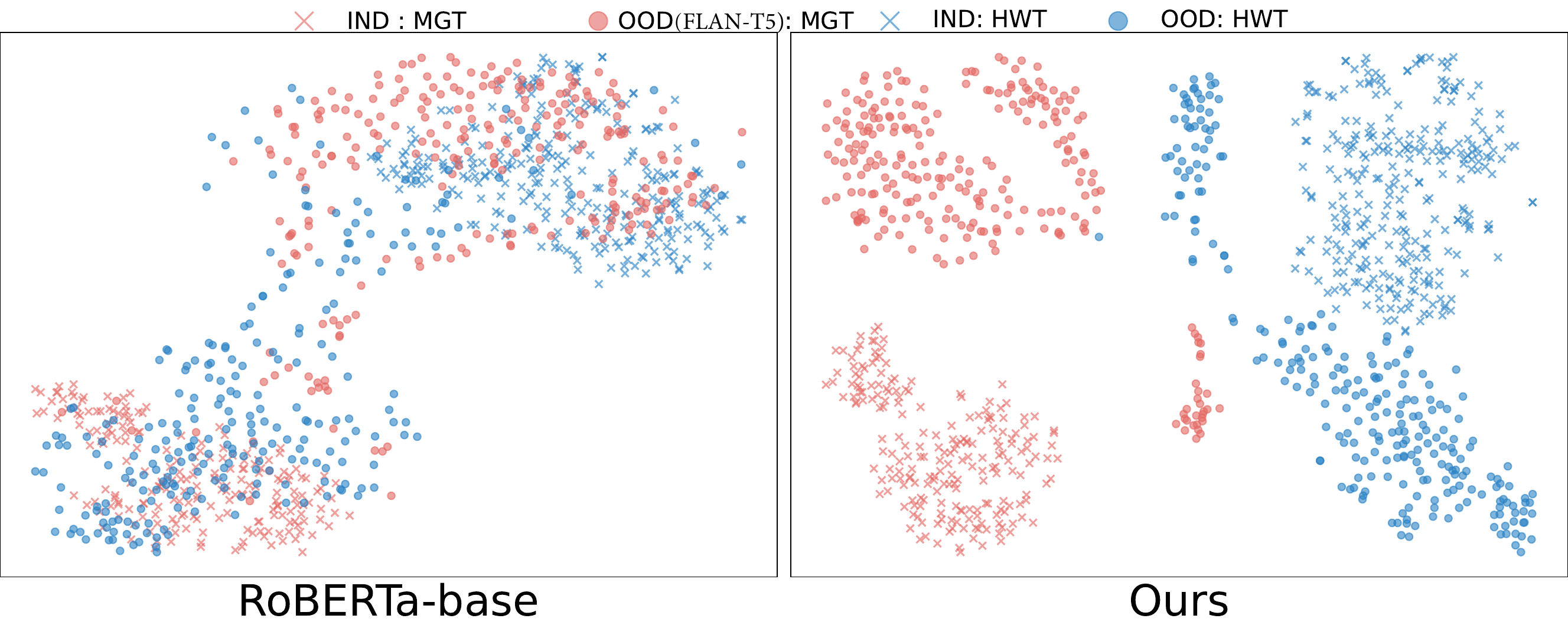}
    \includegraphics[width=\textwidth]{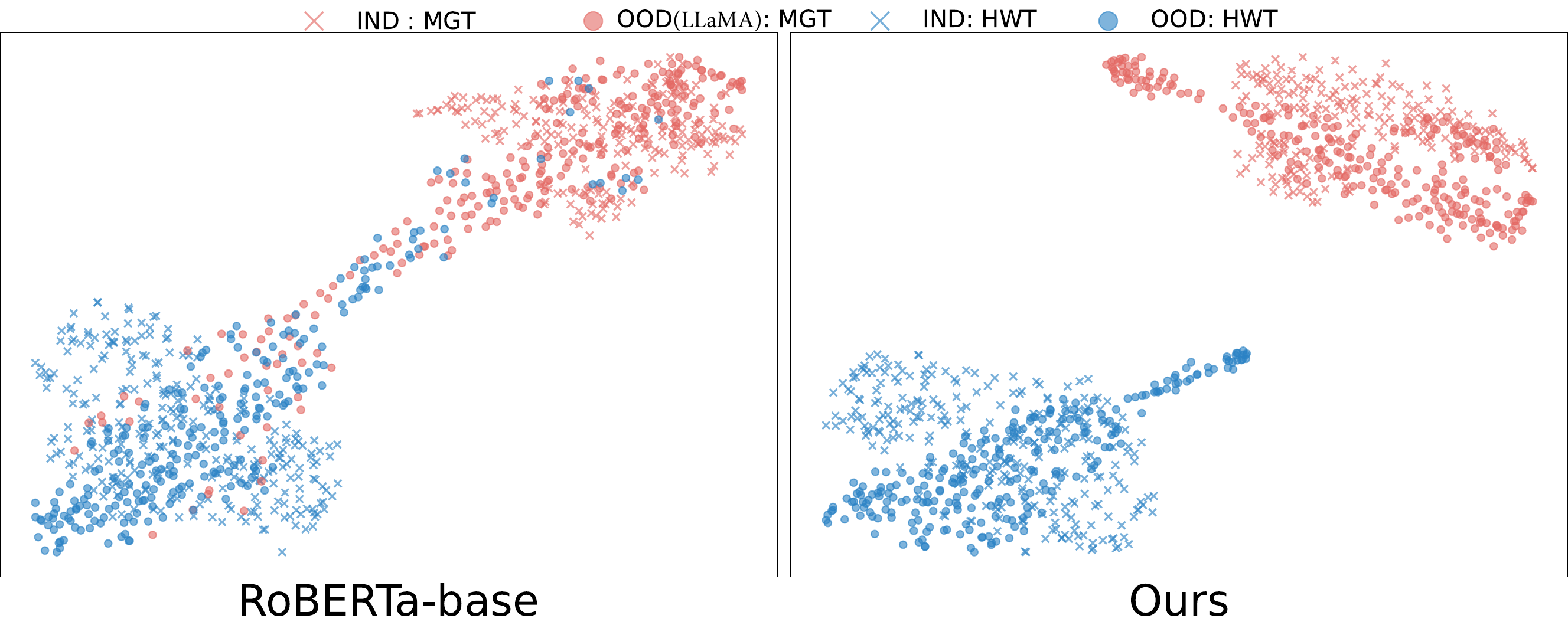}
    \includegraphics[width=\textwidth]{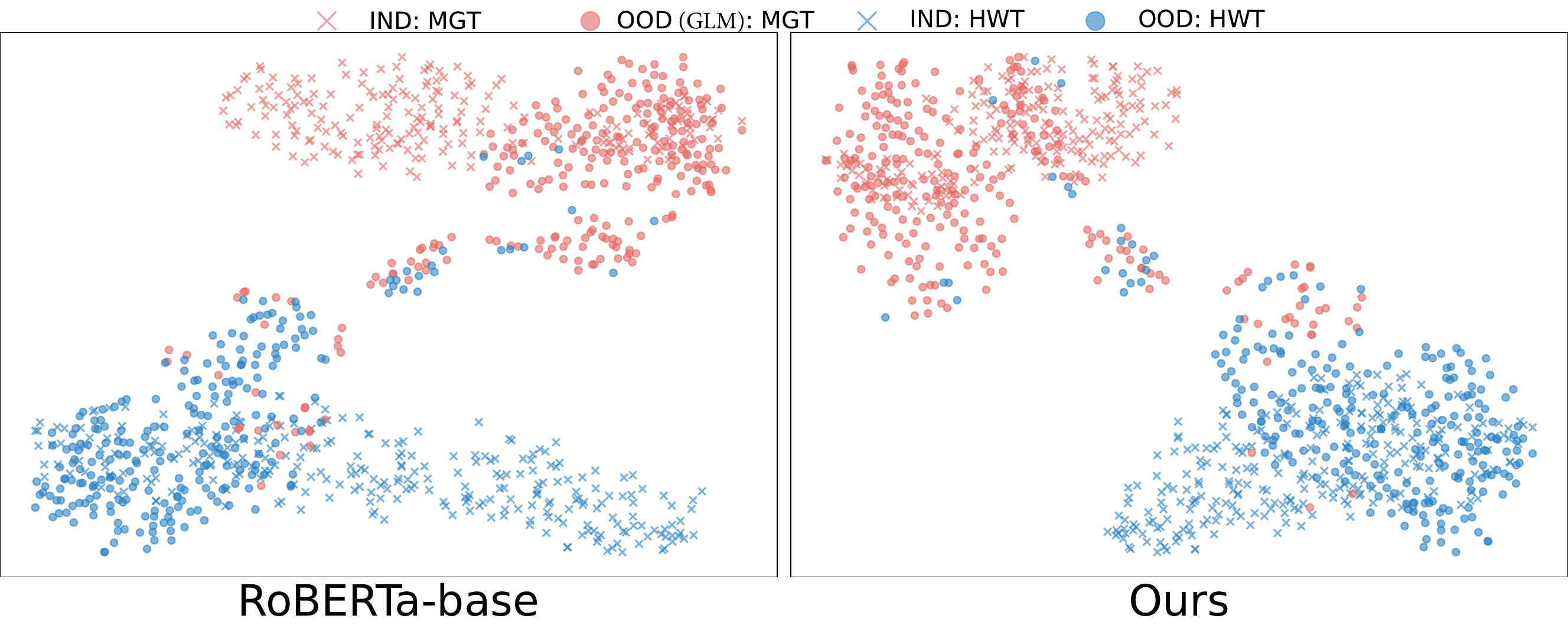}
    \caption{t-SNE visualization of the learned [CLS] embeddings under the Cross-Generator setting.}
    \label{fig:Generator}
\end{figure*}

\begin{figure*}[t]
    \centering
    \includegraphics[width=\textwidth]{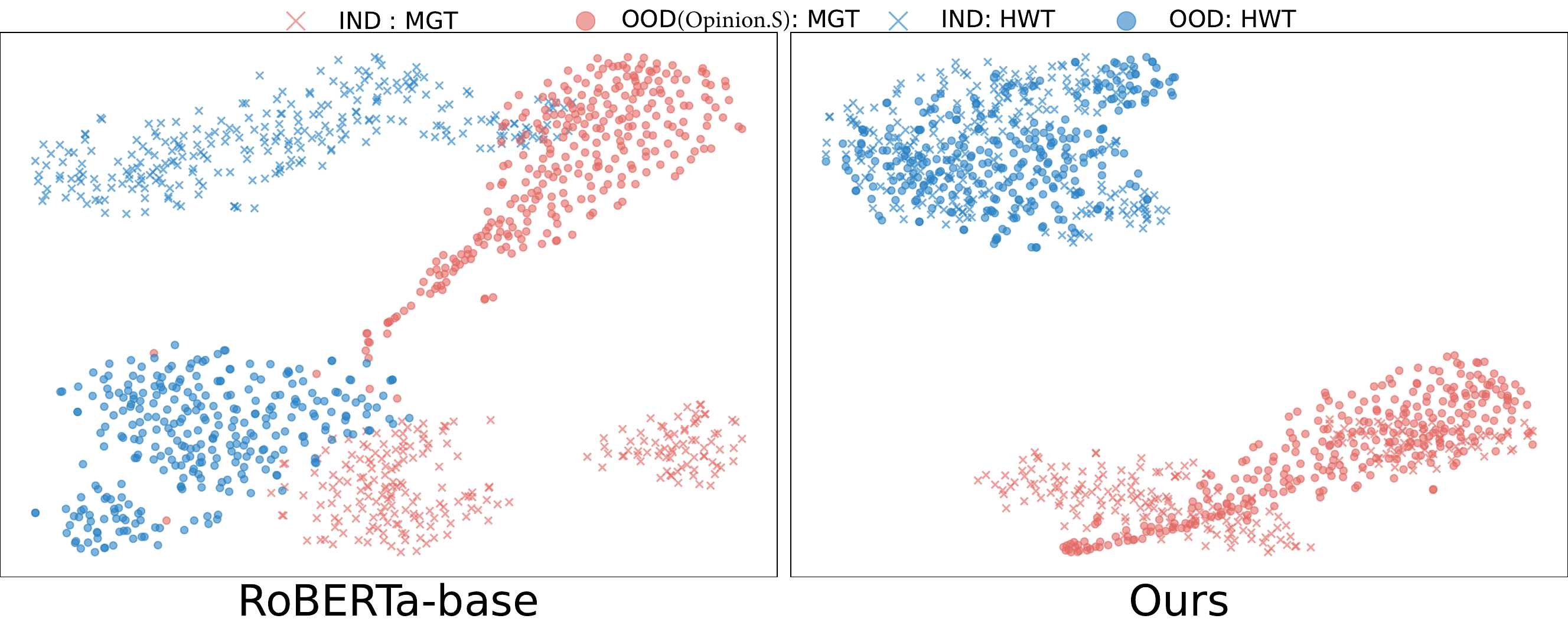}
    \includegraphics[width=\textwidth]{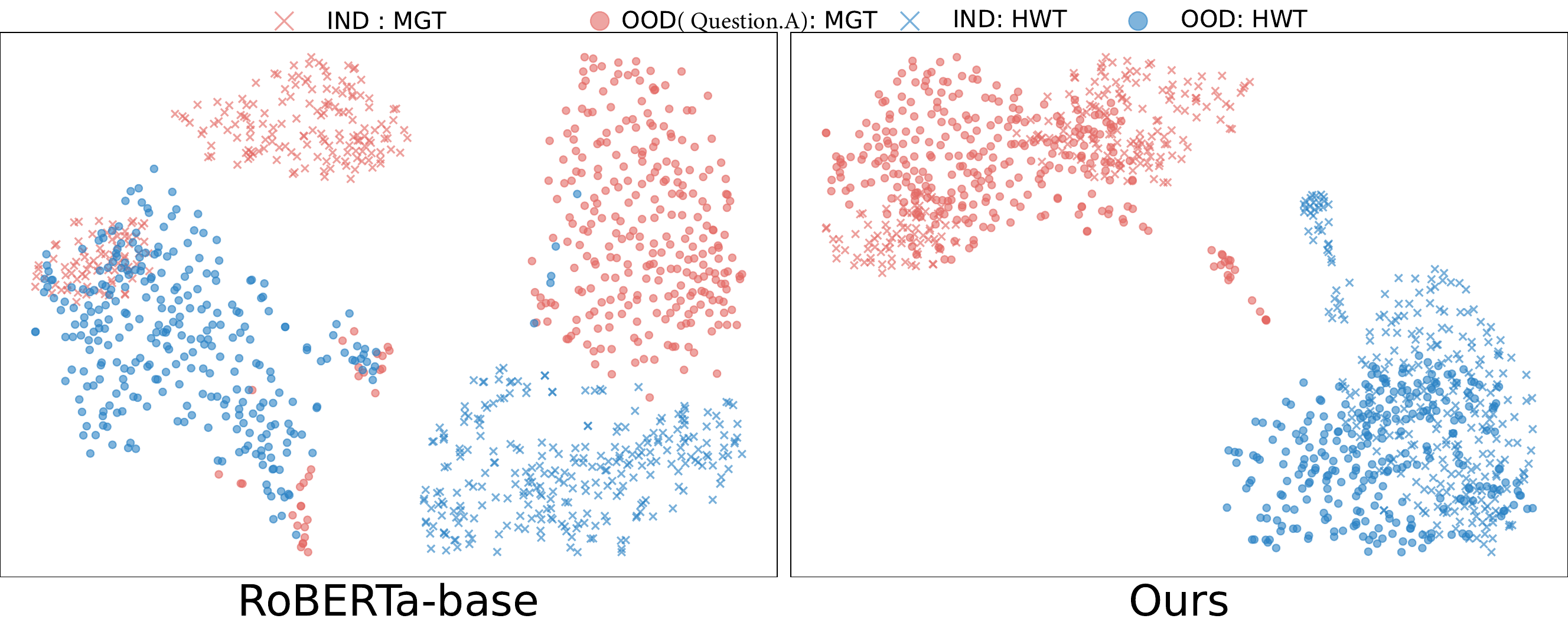}
    \includegraphics[width=\textwidth]{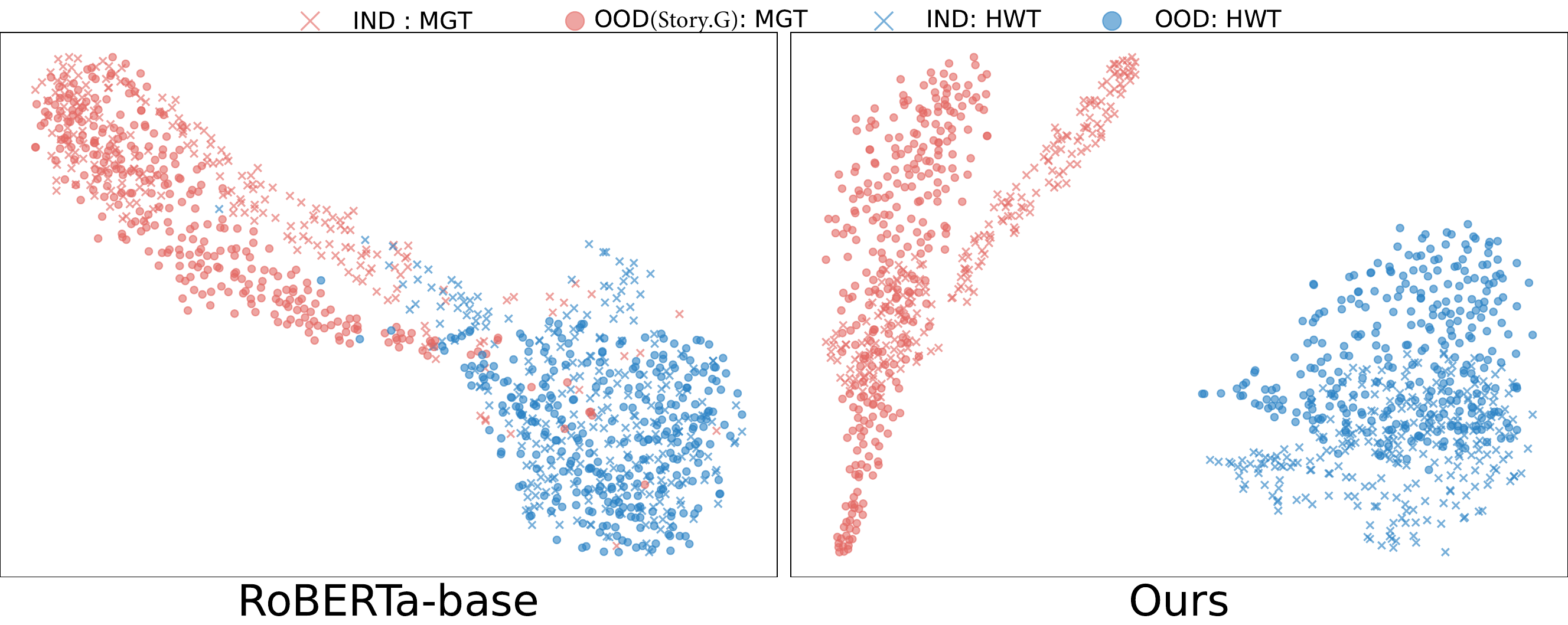}
    \caption{t-SNE visualization of the learned [CLS] embeddings under the Cross-Domain setting.}
    \label{fig:domain}
\end{figure*}

\begin{figure*}[t]
    \centering
    \includegraphics[width=\textwidth]{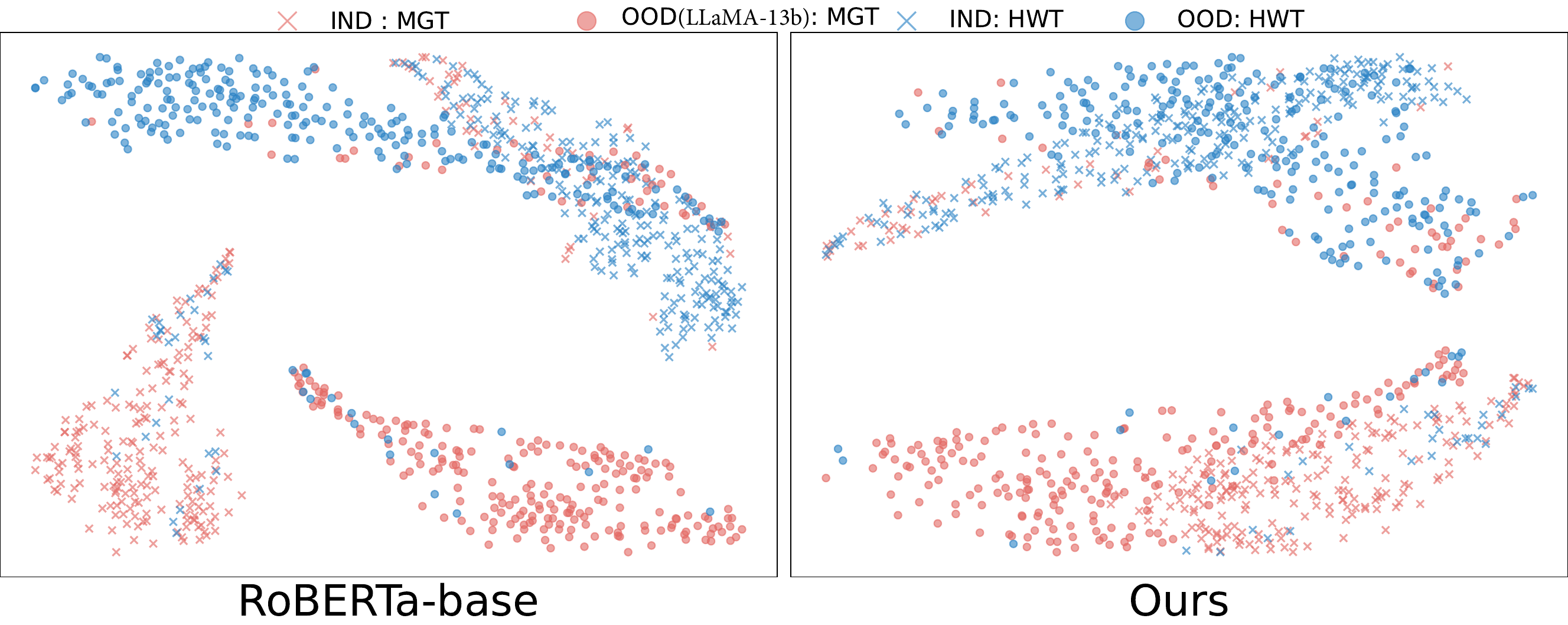}
    \includegraphics[width=\textwidth]{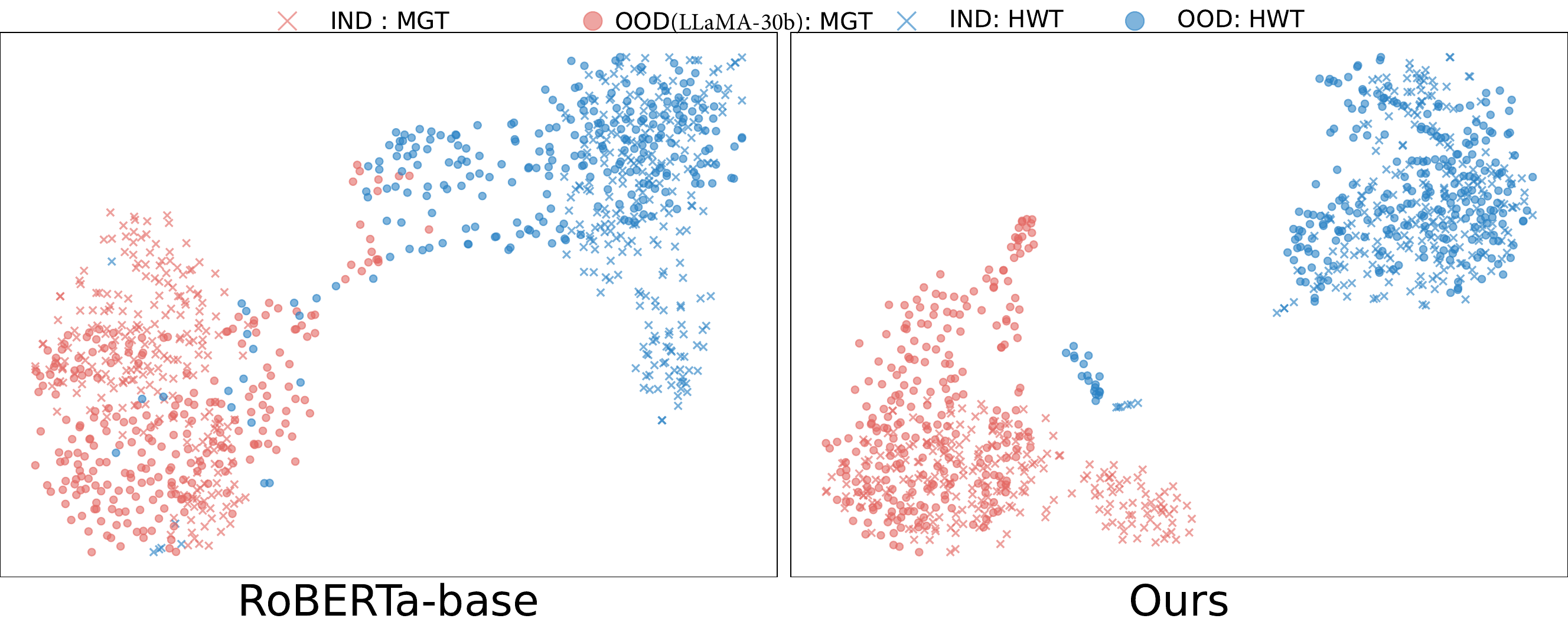}
    \includegraphics[width=\textwidth]{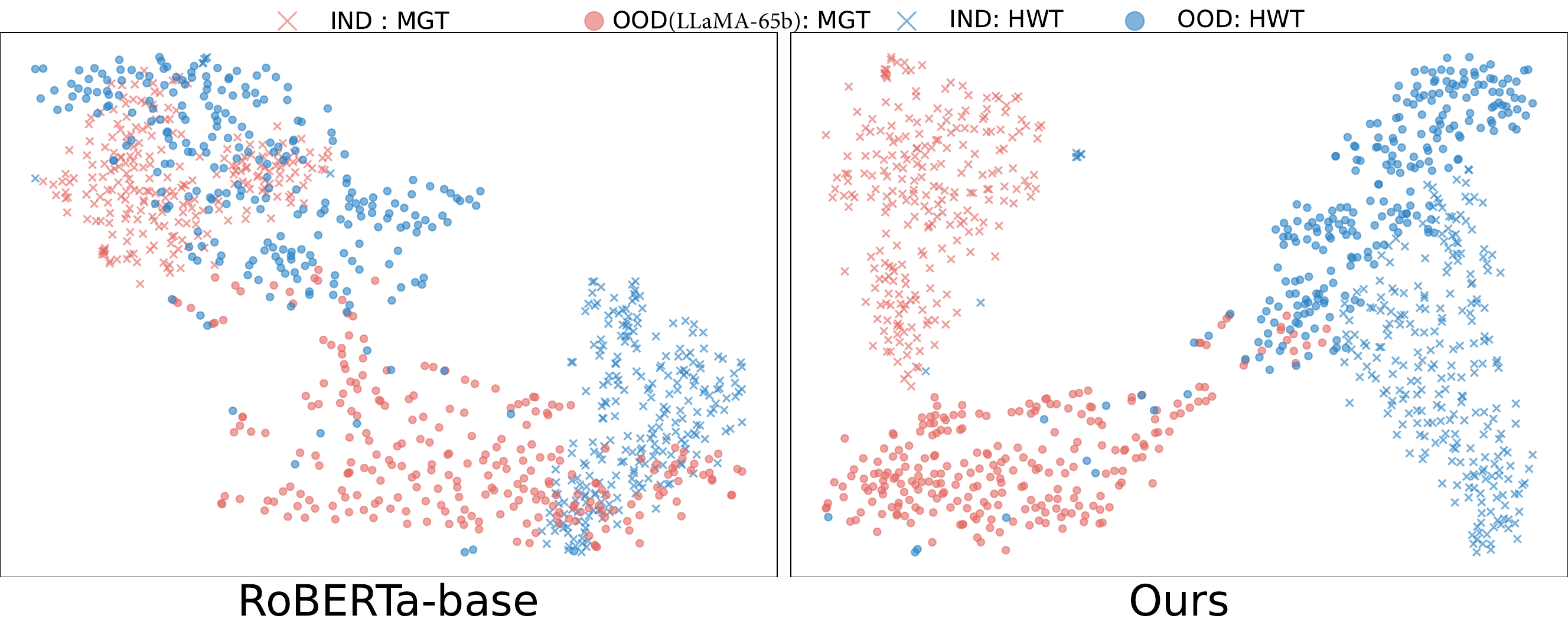}
    \caption{t-SNE visualization of the learned [CLS] embeddings under the Cross-Scale setting.}
    \label{fig:scale}
\end{figure*}

To better understand the improved generalization performance of our method in detecting MGT under domain shifts, we extract features of the test samples from the final hidden layer of both RoBERTa-base and \modelname{}, and use t-SNE~\cite{van2008visualizing} to project them into a 2D space for visualization.
Results in \figref{fig:Generator} to \figref{fig:scale} show the t-SNE visualizations of test sets under three DG scenarios.
From the results, we observe that training and test samples are effectively separated across different test domains. Furthermore, samples with the same label tend to form tighter clusters, which cross-validates our main findings and conclusions.



\end{document}